\documentclass[final,twocolumn]{elsarticle}
\makeatletter
\def\ps@pprintTitle{%
 \let\@oddhead\@empty
 \let\@evenhead\@empty
 \def\@oddfoot{Published in Pattern Recognition - Elsevier: \url{https://doi.org/10.1016/j.patcog.2022.108985}\,\,\,\,\,\,\,\,\,\,\,\,\,\,\,\,\,\,\,\,\,\,\,\,\,\,\,\,\,\,\,\,\,\,\,\,\,\,\,\,\,\,\,\,\,\,\,\,\,\,\,\,\,\,\,\,\,\,\,\,\,\,\,\,}%
 \let\@evenfoot\@oddfoot}
\makeatother

\usepackage{algorithm}
\usepackage{algpseudocode}
\usepackage{bbm}
\usepackage{tabularx}
\usepackage{framed,enumitem} 
\usepackage[hang,flushmargin]{footmisc}
\usepackage{graphicx}
\usepackage{array}
\usepackage{float}
\usepackage{textcomp}
\usepackage{bmpsize}
\usepackage[table]{xcolor}
\usepackage{lipsum}
\usepackage{hyperref}
\usepackage{multirow}
\usepackage{ragged2e}
\usepackage{flushend}
\usepackage{soul}
\usepackage{wrapfig}
\usepackage[export]{adjustbox}
\usepackage{amsmath,amsfonts} %
\usepackage{pifont}%
\newcommand{\xmark}{\ding{55}}%
\newcommand{\greencheck}{{\color{blue}\checkmark}}
\newcommand{\redcross}{{\color{red}\xmark}}
\DeclareMathOperator*{\argmax}{arg\,max}

\newcolumntype{L}[1]{>{\raggedright\let\newline\\\arraybackslash\hspace{0pt}}m{#1}}
\newcolumntype{C}[1]{>{\centering\let\newline\\\arraybackslash\hspace{0pt}}m{#1}}
\newcolumntype{R}[1]{>{\raggedleft\let\newline\\\arraybackslash\hspace{0pt}}m{#1}}
\newcolumntype{Y}{>{\centering\arraybackslash}X}

\definecolor{goodgreen}{RGB}{46, 125, 50}
\definecolor{badred}{RGB}{191, 54, 12}

\definecolor{lorange}{RGB}{255, 243, 224}
\definecolor{gray1}{RGB}{250, 250, 250}
\definecolor{gray2}{RGB}{245, 245, 245}
\definecolor{bluegray}{RGB}{38, 50, 56}

\definecolor{teal}{RGB}{0, 121, 107}
\algnewcommand{\LeftComment}[1]{\Statex \(\triangleright\) #1}
\algnewcommand{\LeftColComment}[1]{\Statex \(\triangleright\) \textcolor{teal}{ #1}}

\begin{document}
\author[a1]{Loris Giulivi\corref{cor1}}
\author[a2]{Malhar Jere}
\author[a1]{Loris Rossi}
\author[a2]{Farinaz Koushanfar}
\author[a3]{Gabriela Ciocarlie}
\author[a4]{Briland Hitaj}
\author[a1]{Giacomo Boracchi}

\address[a1]{Politecnico di Milano, Piazza Leonardo da Vinci, 32,
20133 Milano, Italy}
\address[a2]{University of California San Diego, 9500 Gilman Drive, La Jolla, CA 92093-0021, USA}
\address[a3]{University of Texas San Antonio, One UTSA Circle, San Antonio, TX 78249, USA}
\address[a4]{SRI International, 333 Ravenswood Ave Menlo Park, CA 94025, USA}

\cortext[cor1]{Corresponding author: Via G. Ponzio, 34/5, 20133 Milano, Italy, loris.giulivi@polimi.it}

\title{Adversarial Scratches: Deployable Attacks to CNN Classifiers}

\begin{abstract}
A growing body of work has shown that deep neural networks are susceptible to adversarial examples. These take the form of small perturbations applied to the model's input which lead to incorrect predictions. Unfortunately, most literature focuses on visually imperceivable perturbations to be applied to \emph{digital images} that often are, by design, impossible to be deployed to physical targets. 

We present Adversarial Scratches: a novel $L_0$ black-box attack, which takes the form of scratches in images, and which possesses much greater \emph{deployability} than other state-of-the-art attacks.  Adversarial Scratches leverage Bézier Curves to reduce the dimension of the search space and possibly constrain the attack to a specific location.  

We test Adversarial Scratches in several scenarios, including a publicly available API and images of traffic signs. Results show that our attack achieves higher fooling rate than other deployable state-of-the-art methods, while requiring significantly fewer queries and modifying very few pixels. 

\end{abstract}


\maketitle

\vspace{-2cm}
\vspace{2cm}

\section{Introduction}
\label{sec:intro}

\begin{figure*}[t!]
    \centering
    \renewcommand\tabcolsep{4pt}
    \resizebox{0.9\textwidth}{!}{
        \begin{tabular}{|p{12pt}|p{8pt}p{0.4\textwidth}p{0.4\textwidth}p{2pt}|}
            \multicolumn{4}{c}{\large Deployable and non-deployable perturbations} \vspace{0.15cm} \\
            \hline
            \multirow{7}{*}{\rotatebox[origin=c]{90}{\boldmath$L_1$, $L_2$, $L_\infty$\unboldmath}} & & \includegraphics[width=0.4\textwidth,valign=T]{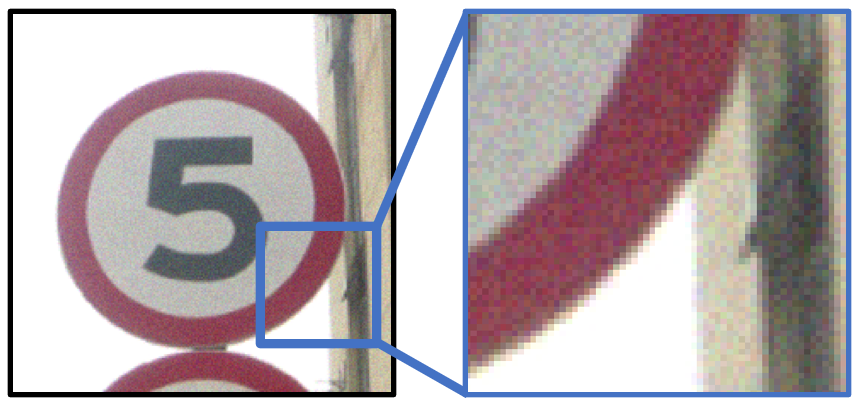}%
            \hspace{0.4cm} \vspace{0.15cm} &
            \vspace{0pt} $L_1$, $L_2$, and $L_\infty$ attacks are not deployable as they potentially modify all the image's pixels. &
            \\
            \hline
            \multirow{27}{*}{\rotatebox[origin=c]{90}{\boldmath$L_0$\unboldmath}} & 
            \multirow{7}{*}{\rotatebox[origin=c]{90}{\textbf{Frame}}} & \includegraphics[width=0.4\textwidth,valign=T]{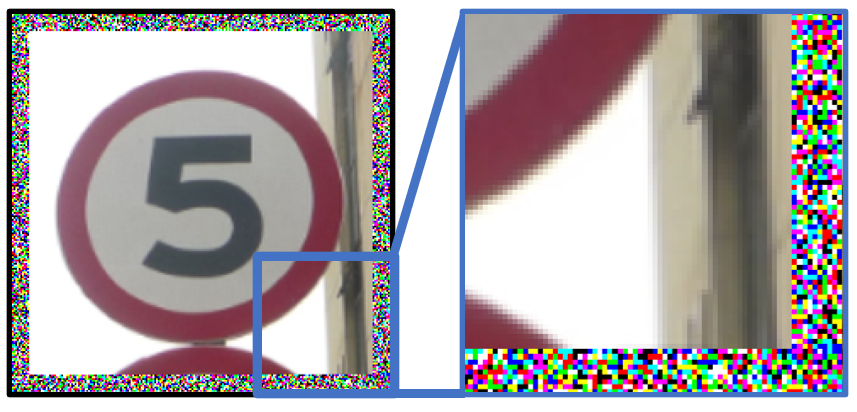}%
            \hspace{0.4cm} \vspace{0.15cm} &
            \vspace{0pt} Frame attacks are not deployable as they modify pixels outside of the target object. &
            \\
            \cline{2-5}
            & \multirow{7}{*}{\rotatebox[origin=c]{90}{\textbf{Sparse}}} &
            \includegraphics[width=0.4\textwidth,valign=T]{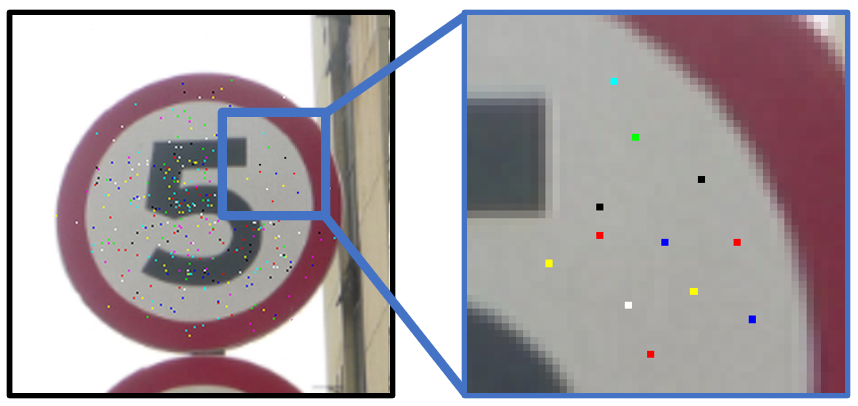}%
            \hspace{0.4cm} \vspace{0.15cm} &
            \vspace{0pt} Sparse attacks, even if localized to the target, are not deployable as they modify several regions in the image. It is unfeasible to apply this perturbation. &
            \vspace{0.15cm}
            \\
            \cline{2-5}
            & \multirow{7}{*}{\rotatebox[origin=c]{90}{\textbf{Patch}}} &
            \includegraphics[width=0.4\textwidth,valign=T]{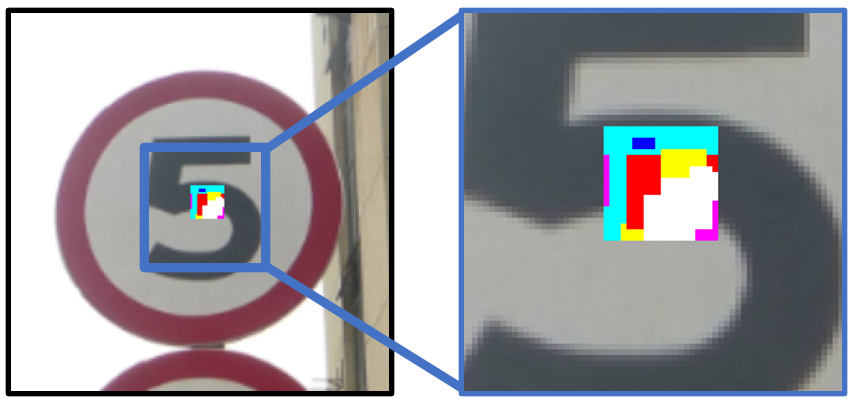}%
            \hspace{0.4cm} &
            \vspace{0pt} The Patch and Scratch attacks are deployable. The perturbations affects spatially contiguous regions, which are entirely contained in the target. &
            \\
            & \multirow{7}{*}{\rotatebox[origin=c]{90}{\textbf{Scratch}}} &
            \includegraphics[width=0.4\textwidth,valign=T]{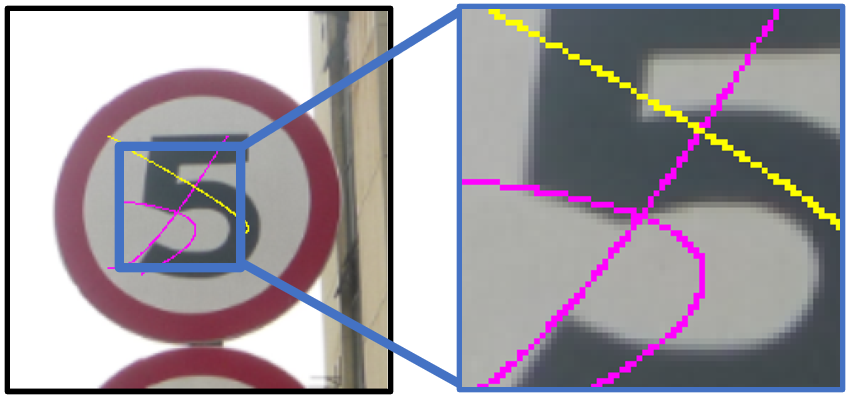}%
            \hspace{0.4cm} \vspace{0.15cm} & &
            \vspace{0pt} 
            \\
            \hline
        \end{tabular}
    }

  \caption{Example adversarial attacks on the TSRD traffic sign dataset. Most perturbations designed for digital images cannot be applied to a physical target. In our work, we focus on deployable perturbations.}
  \label{fig:teasertsrd}
\end{figure*}

Convolutional Neural Networks (CNN)~\cite{lecun1999object} have achieved state-of-the-art performance on a wide array of tasks. These models, however, are surprisingly susceptible to deception by adversarial examples~\cite{szegedy2013intriguing}, consisting in small perturbations to the input that lead to incorrect predictions. The relevance of this security issue is made clear by the increased number of critical systems that make use of CNNs. 
A plethora of papers have explored adversarial vulnerabilities in neural networks, such as~\cite{croce2020sparse} and~\cite{narodytska2017simple}, giving rise to a large corpus of attacks. Typically, attacks are composed of two key components: the perturbation model and the search strategy. 
The former relates to how the image is modified (e.g. the attack only affects a square patch), the latter regards how a successful perturbation is computed, typically requiring an iterative optimization procedure.
Most of the literature, however, focuses on attacks designed for \emph{digital images}. As such, perturbations are confined to the digital domain, and are impossible to be deployed to a physical target. 

Recently, there has been a surging research interest on attacks that can be deployed on real-world systems~\cite{evtimov2020security}. Following this line, in our work we present attacks whose perturbations are designed to be deployable.
We define attacks to be \emph{deployable} based on two required characteristics:
\begin{itemize}[itemindent=0em]
  \item[C1:]  the perturbation only affects the image's pixels that belong to a specific target object;
  \item[C2:] the perturbation is contained in a small, spatially adjacent region.
\end{itemize}
When an attack does not meet the first condition, it means that the perturbation spans multiple objects in the image, possibly including the background. In this case, the attack cannot be applied to a physical target (for example, $L1, L2, L_\infty$ and Frame attacks in Figure \ref{fig:teasertsrd}). When the second condition is not met, the attack may modify a multitude of regions in the target. Clearly, accurately accounting for the relative position of these regions on the target is hardly feasible, rendering the attack non-deployable (for example, the Sparse attack in Figure \ref{fig:teasertsrd}).
Attacks such as~\cite{croce2020sparse, yang2020patchattack} are some of the rare examples that satisfy both (C1) and (C2), and rely on localizing the perturbation inside a square patch (Patch attack in figure \ref{fig:teasertsrd}). However, these either require many attempts (high query requirement)~\cite{croce2020sparse} or large patches~\cite{yang2020patchattack} to be successful.

\begin{figure*}[t]
    \centering
    \renewcommand\tabcolsep{0pt}
    \resizebox{1.0\textwidth}{!}{
        \begin{tabularx}{\textwidth}{p{8pt}YYYYp{10pt}p{14pt}}
            ~ &\multicolumn{4}{c}{\vspace{0pt} \includegraphics[width=0.90\textwidth,valign=T]{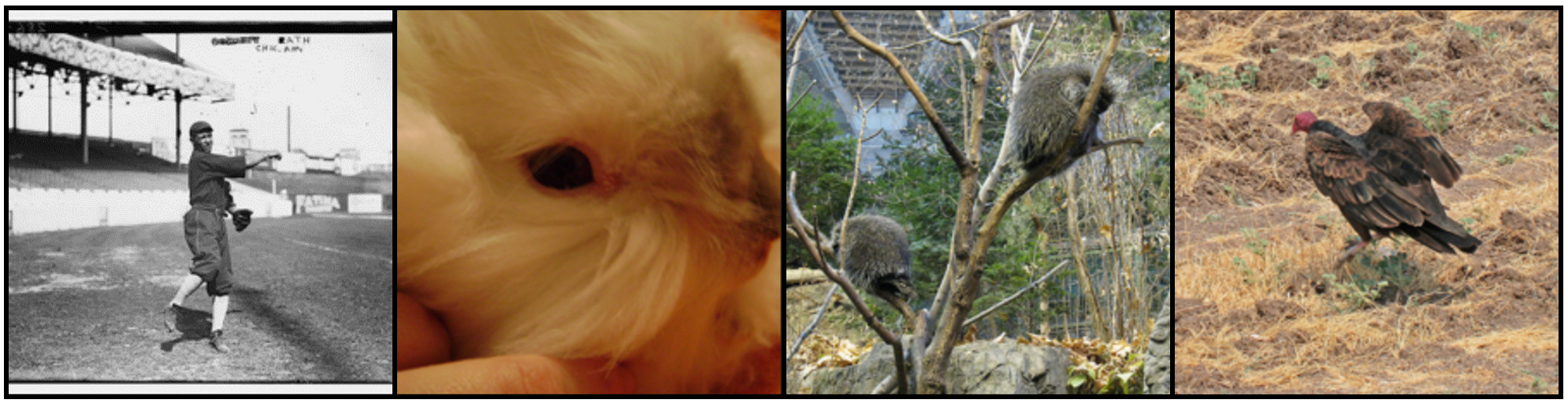}} & ~ & \multirow{8}[8]{*}{\rotatebox[origin=c]{90}{\textbf{\,~Original}}} \\
            ~ & Baseball player & Angora rabbit & Hedgehog & Vulture & ~ & ~ \\
            ~ &\multicolumn{4}{c}{\includegraphics[width=0.90\textwidth,valign=T]{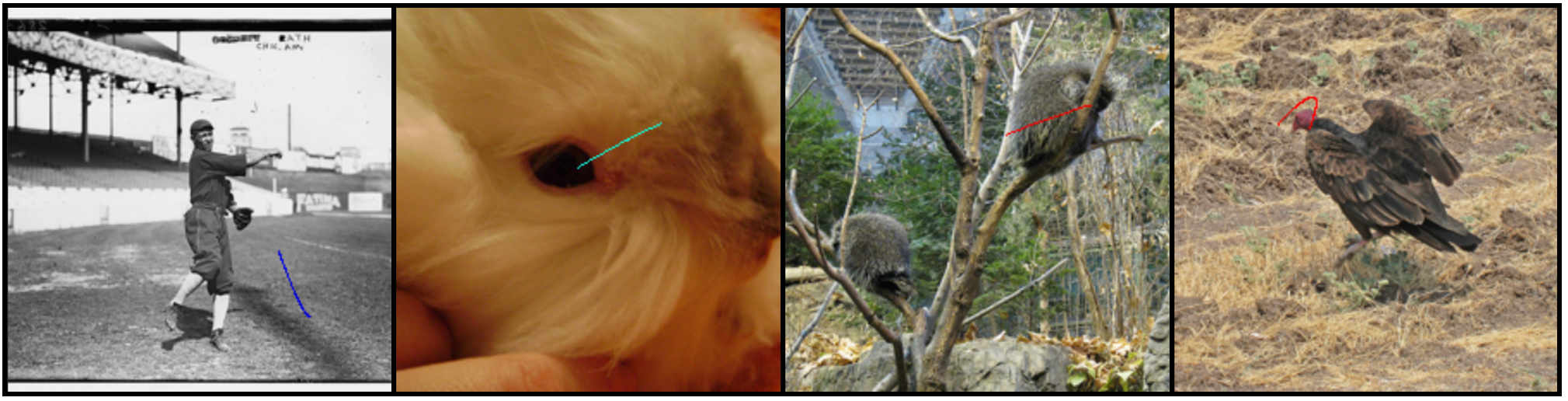}} & ~ & \multirow{8}[8]{*}{\rotatebox{90}{\textbf{Adversarial}}}\\
            ~ & Ski & Maltese dog & Baboon & Hornbill & ~ & ~ \\
        \end{tabularx}
    }
    \caption{Successful Adversarial Scratches on ImageNet (one scratch, $L_0=50$). Adversarial Scratches are powerful attacks that both require a minimal perturbation (small $L_0$) and a low number of queries to be successful. On top of this, the adjacency of the perturbed region enables this attack to be deployable in the physical world.}
    \label{fig:teaser}
\end{figure*}

In this work, we propose \emph{Adversarial Scratches}: a novel, powerful, deployable attack, illustrated in the Scratch attack in Figure \ref{fig:teasertsrd} and in Figure \ref{fig:teaser}.
Adversarial Scratches are constituted of parametric curves, resembling graffiti or small damages when applied to a target, and are thus spatially adjacent~(C2). In particular, we make use of Bézier curves, as these can express a wide variety of shapes. Our intuition is that these curves may introduce patterns in the image to which CNNs are sensitive. Moreover, their compact parametric representation allows efficient optimization. Crucially, Bézier curves can be arbitrarily clipped (see Section \ref{sec:scratchclipping}), allowing to confine the attack to the target region~(C1).
Lastly, Adversarial Scratches are set in the black-box attack scenario, meaning that the model's internals are not known when computing the attack. As such, Adversarial Scratches leverage a variety of gradient-free search strategies.
Our contributions are the following:

\begin{itemize}[leftmargin=*]
    \item \textit{\textbf{Adversarial Scratches:}} We introduce a new perturbation model which is deployable by design. Adversarial Scratches only perturb small regions in the image in the shape of Bézier curves. Furthermore, the parametric nature of Adversarial Scratches enables to achieve state-of-the-art fooling rates while greatly reducing query requirements.
    
    \item \textit{\textbf{Countermeasures:}} We propose two countermeasures to our attacks, namely, median filtering and JPEG compression, and assess their impact in mitigating the effects of Adversarial Scratches.
    
    \item \textit{\textbf{Adversarial software framework:}} We release a Python open-source tool to design and perform adversarial attacks on several popular CNNs for image classification. Within our framework, we also provide interfaces for various optimization strategies, perturbation models, and target networks.
    
    \item \textit{\textbf{A benchmark for deployable attacks:}} 
    We manually segment target regions for a subset of samples of the TSRD \cite{TSRD} dataset, and make these segmentations publicly available. The improved dataset poses as a benchmark for simulation of deplyable attacks on \emph{in-the-wild} traffic signs.
\end{itemize}

\noindent Our code and datasets are made publicly available \href{https://github.com/loris2222/AdversarialScratches/}{\texttt{here}}.

Our extensive experimental evaluation analyzes the performance of Adversarial Scratches across a variety of scenarios.
Firstly, we consider a well-established test-bed on ImageNet, and compare against other deployable and non-deployable state-of-the-art attacks. 
Secondly, we design an experiment utilizing our own-developed version of the TSRD dataset, where perturbations are applied to images of traffic signs.
Furthermore, we launch an attack against Microsoft’s commercially available Cognitive Services Image Captioning API. Our attacks successfully fooled the API, demonstrating the effectiveness of Adversarial Scratches on a production-grade Machine-Learning-as-a-service system. \emph{We have contacted Microsoft regarding this vulnerability}.
Lastly, we study the performance of Adversarial Scratches across a variety of optimization strategies and parametric configurations.

\section{Background and Related Work}
\label{sec:relwork}
After providing a formal description of the problem, we categorize adversarial attacks on neural networks and overview the state-of-the-art.

\subsection{Problem Formulation}
\label{sec:problemformulation}
We denote as $\mathbf{x} \in [0,1]^{w,h,c}$ an image having width $w$, height $h$, and $c$ channels, and as $f(\cdot)$ an image classifier (a CNN in our case). We assume the CNN to return a vector $f(\mathbf{x})$ where each component $f(\mathbf{x})_{i}$ represents the posterior probability of $\mathbf{x}$ belonging to class $i$, thus, the label assigned by the CNN to $\mathbf{x}$ is $C(\mathbf{x}) = \argmax_{i} f(\mathbf{x})_{i}$. We denote the ground truth label as $y$.

Given a correctly classified sample $\mathbf{x}$, namely, $C(\mathbf{x}) = y$, the problem consists in finding an adversarial sample $\mathbf{x}'\in [0,1]^{w,h,c}$ which, according to a distance metric $L$ and a distance threshold $\delta$, is close to the original sample $\mathbf{x}$:
\begin{equation}
    L(\mathbf{x}',\mathbf{x}) < \delta \,,
\label{equation:problemconstraint}
\end{equation}
such that \emph{the model is fooled}, which means:
\begin{equation} 
\label{equation:misclassification}
    C(\mathbf{x}') \neq y .
\end{equation}
Typically, the distance $L$ is realized in the $L_p$ norm of the difference between the original and the adversarial samples, namely $L(\mathbf{x}',\mathbf{x}) = \|\mathbf{x}'-\mathbf{x}\|_p$. We consider the $L_0$ norm as this is the only metric enabling spatially contained perturbations, and in turn satisfying (C2) (Figure \ref{fig:teasertsrd}).
Another requirement for deployability of the attack is constraining the perturbation to the target's surface (C1). Therefore, for each image $\mathbf{x}$, we also require a target region $\mathbf{r}$ indicating the pixels of $\mathbf{x}$ that belong to the target object. Figure \ref{fig:attackmodel} shows how this region is used.

\subsection{Prior Works and Categorization of Attacks}
\label{sec:categorization}
Biggio et al.~\cite{biggio2012poisoning, biggio_first_dnn} were the first to present adversarial examples in gradient-based learning systems, such as support vector machines (SVMs) and neural networks. Szegedy et al.~\cite{szegedy2013intriguing} discovered that this issue also extends to ImageNet-trained deep neural networks. Other works show that adversarial examples can target models addressing tasks other than classification, such as clustering~\cite{cina2022black}.
Attacks to neural networks are characterized by two major ingredients:  \emph{i)} the \emph{perturbation model}, which places constraints on the attack, thereby defining the feasible search space for $\mathbf{x}'$, and \emph{ii)} the \emph{search strategy} used to explore this space. 

Perturbation models can be categorized along four axes. The first regards the access level the attacker has on the target neural network (i.e., attack surface), the second specifies how the perturbation magnitude is measured (i.e., metric), the third regards specific constraints on the support of the perturbation (i.e., geometric structure), and the fourth denotes the aim of the attack.

\noindent\textbf{Attack Surface:}
    along this axis, perturbation models are categorized in white-box and black-box. The first category indicates that the attacker has a full view of the internals of the model $f$, including its gradients, the latter instead specifies that the attacker may only control the input $\mathbf{x}'$ and observe its output. Black-box attacks can also be decomposed into \emph{score-based} attacks, when the full class score vector $f(\mathbf{x}')$ is provided \cite{guo2019simple}, or \emph{decision-based} ones \cite{brendel2018decision}, when only the predicted label $C(\mathbf{x}')$ is provided. An example of a white-box attack is the JSMA attack~\cite{papernot2016limitations}, which finds vulnerable pixels through saliency maps. Regarding black-box attacks, we mention DEceit~\cite{ghosh2022black}, which uses differential evolution to optimize an attack with adjustable sparsity.
    
    White-box attacks pose no serious threats to production-level systems, as it is unlikely that providers would disclose information regarding their models. Thus, we choose to frame ourselves in the black-box setting, and allow our search strategy to only control the model input and have access only to its output.

\noindent\textbf{Metric:}
    On the second axis, perturbation models are categorized depending on the particular $L_p$ norm chosen to constrain the magnitude of the applied perturbation, which is measured by $\|\mathbf{x}'-\mathbf{x}\|_p$. $L_{0}$ attacks measure and regulate the number of perturbed pixels, $L_{1}$ and $L_{2}$ attacks the Manhattan and Euclidean distance between the original and the perturbed image, and $L_{\infty}$ attacks the largest pixel-wise difference between the two. An adversarial sample $\mathbf{x}'$ is deemed valid when it is able to fool the model (\ref{equation:misclassification}) and when it is inside the $L_{p}$ ball of a specified radius centered in $\mathbf{x}$ (\ref{equation:problemconstraint}).
    
    Recent literature has proposed attacks in the $L_{1}$, $L_{2}$, and $L_{\infty}$ norm constraint scenarios. Perhaps the most common are $L_{\infty}$ attacks, such as~\cite{moonICML19}, which uses a surrogate problem to modify image patches, and~\cite{li2021universal} which computes universal patches to fool object detectors.
    Typically, $L_1, L_2, $ and $L_\infty$ attacks alter pixels by very small amounts, giving rise to human-imperceptible perturbations. However, by not constraining the number of perturbed pixels, these attacks often modify the large majority of the image (Figure \ref{fig:teasertsrd}, $L_1, L_2, L_\infty$ attacks), resulting in non-deployable perturbations.
    In contrast, attacks of $L_0$ nature such as \cite{narodytska2017simple, croce2020sparse} by definition limit the number of perturbed pixels.
    Thus, we choose to set Adversarial Scratches in the $L_{0}$-bounded scenario. 
    
\noindent\textbf{Perturbation Structure:}
    We categorize attacks on this axis based on specific constraints imposed on the perturbation, such as its localization or geometric structure (e.g., patches~\cite{yang2020patchattack}, objects~\cite{xiao2021fooling}, signatures~\cite{li2021black}). As an example of a structured attack, we mention Patch Attack~\cite{yang2020patchattack}, which uses reinforcement learning to optimally place pre-generated textured patches, albeit covering up to $20\%$ of the entire image.
    Adversarial Scratches are structured as contiguous curves, this being a key element for the deployability of our attack. Indeed, as shown in Fig \ref{fig:teasertsrd}, the perturbation of an unstructured attack, even of the $L_0$ type, cannot be contained in a spatially adjacent region in the image (C2).
    
\noindent\textbf{Aim:}
    We finally sort attacks into targeted ones, for which the goal is to force the prediction to a specific class $y'$, such that $C(\mathbf{x}') = y'$, and untargeted ones, where the objective is simply to induce misclassification, independently of the resulting class, such that $C(\mathbf{x}') \neq y = C(\mathbf{x})$.\
    For this categorization, we cite Sparse-RS~\cite{croce2020sparse}, as it displays both targeted and untargeted attacks.
    In this work, we primarily focus on untargeted attacks. In Section~\ref{sec:targetedattacks}, we discuss how to obtain targeted Adversarial Scratches.

\vspace{0.3cm}

\begin{table*}[t]
    \small
    \renewcommand\arraystretch{1.2}
    \caption{Comparison of methodologies in the literature and our own.}
    \centering
    \resizebox{\textwidth}{!}{
    \begin{tabular}{|c|c|c|c|c|c|c|}
    \hline
    \textbf{Method} & \textbf{Search strategy} & \textbf{\begin{tabular}[c]{@{}c@{}}Perturbation\\  structure\end{tabular}} & \textbf{\begin{tabular}[c]{@{}c@{}}Large\\  Network\\  (ImageNet)\end{tabular}} & \textbf{\begin{tabular}[c]{@{}c@{}}Localized\\ Perturbation\\ (L0 limited)?\end{tabular}} & \textbf{\begin{tabular}[c]{@{}c@{}}Deployable\\ attack?\end{tabular}} \\ \hline
    \begin{tabular}[c]{@{}c@{}}Narodytska and\\  Kasiviswanathan, 2016~\cite{narodytska2017simple}\end{tabular} & local random search & sparse & \greencheck & \greencheck & \redcross \\ \hline
    JSMA~\cite{papernot2016limitations} & \begin{tabular}[c]{@{}c@{}}substitute model\end{tabular} & sparse & \redcross & \greencheck & \redcross \\ \hline
    \begin{tabular}[c]{@{}c@{}}Boundary attack~\cite{brendel2018decision}\end{tabular} & \begin{tabular}[c]{@{}c@{}}gaussian perturbation\end{tabular} & unstructured & \greencheck & \redcross & \redcross \\ \hline
    \begin{tabular}[c]{@{}c@{}}Ilyas et al. 2018~\cite{ilyas2018black}\end{tabular} & \begin{tabular}[c]{@{}c@{}}natural evolution\\  strategy\end{tabular} & unstructured & \greencheck & \redcross & \redcross \\ \hline
    CornerSearch~\cite{croce2019sparse} & \begin{tabular}[c]{@{}c@{}}coordinate-wise\\ gradient estimation\end{tabular} & sparse & \redcross & \greencheck & \redcross \\ \hline
    \begin{tabular}[c]{@{}c@{}}SimBA~\cite{guo2019simple}\end{tabular} & \begin{tabular}[c]{@{}c@{}}Fourier coefficients\end{tabular} & \begin{tabular}[c]{@{}c@{}}unstructured\\  sparse\end{tabular} & \greencheck & \greencheck & \redcross \\ \hline
    \begin{tabular}[c]{@{}c@{}}Parsimonious Black-Box\\ Adversarial Attacks~\cite{moonICML19}\end{tabular} & \begin{tabular}[c]{@{}c@{}}greedy\\  local search\end{tabular} & unstructured & \greencheck & \redcross & \redcross \\ \hline
    PatchAttack~\cite{yang2020patchattack} & \begin{tabular}[c]{@{}c@{}}RL  agent\end{tabular} & patch & \greencheck & \greencheck & \greencheck \\ \hline
    DEceit~\cite{ghosh2022black} & \begin{tabular}[c]{@{}c@{}}differential evolution\end{tabular} & sparse & \greencheck & \greencheck & \redcross\\ \hline
    Sparse-RS~\cite{croce2020sparse} (``any-pixel'') & \begin{tabular}[c]{@{}c@{}}random search\end{tabular} & sparse & \greencheck & \greencheck & \redcross\\ \hline
    Sparse-RS~\cite{croce2020sparse} (Patch-RS) & \begin{tabular}[c]{@{}c@{}}random search\end{tabular} & patch & \greencheck & \greencheck & \greencheck\\ \hline
    \textbf{Adversarial Scratches} (Ours) & \begin{tabular}[c]{@{}c@{}}NGO \end{tabular} & scratch & \greencheck & \greencheck & \greencheck \\ \hline
    \end{tabular}
    }
    \label{tab:attack_comparisons}
\end{table*}

In summary, Adversarial Scratches belong to the black-box (score-based) category, are $L_0$-bounded, and adopt the structure of deployable scratches.
Alongside this formal categorization, we report in Table~\ref{tab:attack_comparisons} a comparison of popular perturbation models in the literature, where we consider a variety of different factors, such as whether the attacks are $L_0$-bounded, were launched on a large, ImageNet-scale network, and most importantly whether these were deployable. The analysis shows strong similarities between our work and Patch-RS~\cite{croce2020sparse}, the primary difference being the perturbation model.
The attacks presented in Sparse-RS achieve state-of-the-art performance in several settings. In particular, Patch-RS is, to the best of our knowledge, the most effective black-box $L_0$ attack which also allows deployable perturbations, by modifying a single square patch in the image. As such, this is the most relevant method amongst those considered in our experiments. Nonetheless, we also compare to non-deployable attacks, since they represent an ideal reference amongst all $L_0$ attacks. The ``any-pixel" attack from~\cite{croce2020sparse}, which modifies any $k$ pixels in the image, shows the best results amongst non-deployable $L_0$ attacks. We remark that comparing deployable attacks to non-deployable attacks is unfair, as the latter can exploit a much larger attack surface.

\section{Methodology}
We frame the generation of Adversarial Scratches as a constrained optimization problem. Given a trained CNN classifier $f$ and an input image $\mathbf{x}$, the adversarial sample $\mathbf{x}'$ is found by minimizing the \emph{margin loss}:
\begin{equation}
\label{eq:margin_loss}
    \mathcal{L}_f(\mathbf{x}, \mathbf{x}') = f(\mathbf{x}')_{y} - \max_{i\neq y}(f(\mathbf{x}')_i)\,
\end{equation}
subject to the $L_0$ bound:
\begin{equation}
    \label{equation:l0_constraint}
    \|\mathbf{x}'-\mathbf{x}\|_0 \leq k,
\end{equation}
and to the localization constraint:
\begin{equation} 
    \label{equation:localization_constraint}
    (\mathbf{r}[i,j]=0) \implies (\mathbf{x'}[i,j] = \mathbf{x}[i,j]).
\end{equation}
We define the margin loss (\ref{eq:margin_loss}) $\mathcal{L}_f$ as in~\cite{croce2020sparse} as the difference between the posterior probability $f(\mathbf{x}')_{y}$ of $\mathbf{x}'$ belonging to the ground truth class $y$, and the maximum posterior probability $\max_{i\neq y}(f(\mathbf{x}')_i)$ of $\mathbf{x}'$ belonging to a class other than $y$.
The sample $\mathbf{x}'$ is misclassified when $\max_{i\neq y}(f(\mathbf{x}')_i) > f(\mathbf{x}')_{y}$.\footnote{This holds also when $\mathbf{x}$ is not correctly classified ($C(\mathbf{x}) \neq y$), and in this case we have $\mathcal{L}_f(\mathbf{x},\mathbf{x}) < 0$. However, $\mathbf{x}$ can never be considered as an adversarial sample for itself. For this reason, in our experiments, we follow the common practice of discarding all misclassified images $\mathbf{x}$.} Therefore, $\mathbf{x}'$ is an adversarial sample for $\mathbf{x}$ when:
\begin{equation}
\label{eq:margin_loss_target}
    \mathcal{L}_f(\mathbf{x}, \mathbf{x}') < 0 .
\end{equation}

We solve the constrained optimization problem through an iterative optimization procedure. The result is an image $\mathbf{x}^v$ obtained by superimposing a scratch $B$ to the original sample $\mathbf{x}$, where $B$ is modeled as a Bézier curve identified by parameter vector $v$.  Figure \ref{fig:attackmodel} visualizes how the perturbation is computed starting from this parameter configuration, highlighting the usage of region $\mathbf{r}$. In the remainder of the section, we detail the perturbation model and the search strategy that characterize Adversarial Scratches.

\subsection{Perturbation Model}
\label{sec:AdversarialScratches}
\begin{table*}[t!]
    \centering
    \caption{Description of quadratic Bézier curve parametrization}
    \label{tab:parametrization}
    \resizebox{1.0\textwidth}{!}{
    \begin{tabular}{|c|c|c|c|c|}
        \hline
        Parameter & Interpretation & Min value & Max value & Description  \\
        \hline
        $v_0$ & $P_{0,x}$ & 0 & $w-1$ & $x$ coordinate of point $P_0$  \\
        $v_1$ & $P_{0,y}$ & 0 & $h-1$ & $y$ coordinate of point $P_0$  \\
        $v_2$ & $P_{1,x}$ & 0 & $w-1$ & $x$ coordinate of point $P_1$  \\
        $v_3$ & $P_{1,y}$ & 0 & $h-1$ & $y$ coordinate of point $P_1$  \\
        $v_4$ & $P_{2,x}$ & 0 & $w-1$ & $x$ coordinate of point $P_2$  \\
        $v_5$ & $P_{2,y}$ & 0 & $h-1$ & $y$ coordinate of point $P_2$  \\
        $v_6$ & $R$ & 0 & 255 & red component of scratch color*  \\
        $v_7$ & $G$ & 0 & 255 & green component of scratch color*  \\
        $v_8$ & $B$ & 0 & 255 & blue component of scratch color*  \\
        \hline
    \end{tabular}
    }
    \\
    \begin{flushleft}
    \rightskip=0pt \justifying * Colors are restricted to be fully saturated, thus, the color components can only assume the extreme values $0$ or $255$ of the range. This means that there are only eight available colors: [0, 0, 0], [255, 0, 0], [0, 255, 0], [0, 0, 255], [255, 255, 0], [0, 255, 255], [255, 0, 255], and [255, 255, 255].
    \end{flushleft}
    \vspace{-15pt}
\end{table*}

\begin{figure*}[t]
    \centering
    \resizebox{1.0\textwidth}{!}{
        \begin{picture}(1000,498)
            \fontsize{100}{120}
            \selectfont
            \put(0,30){\includegraphics[width=1000pt]{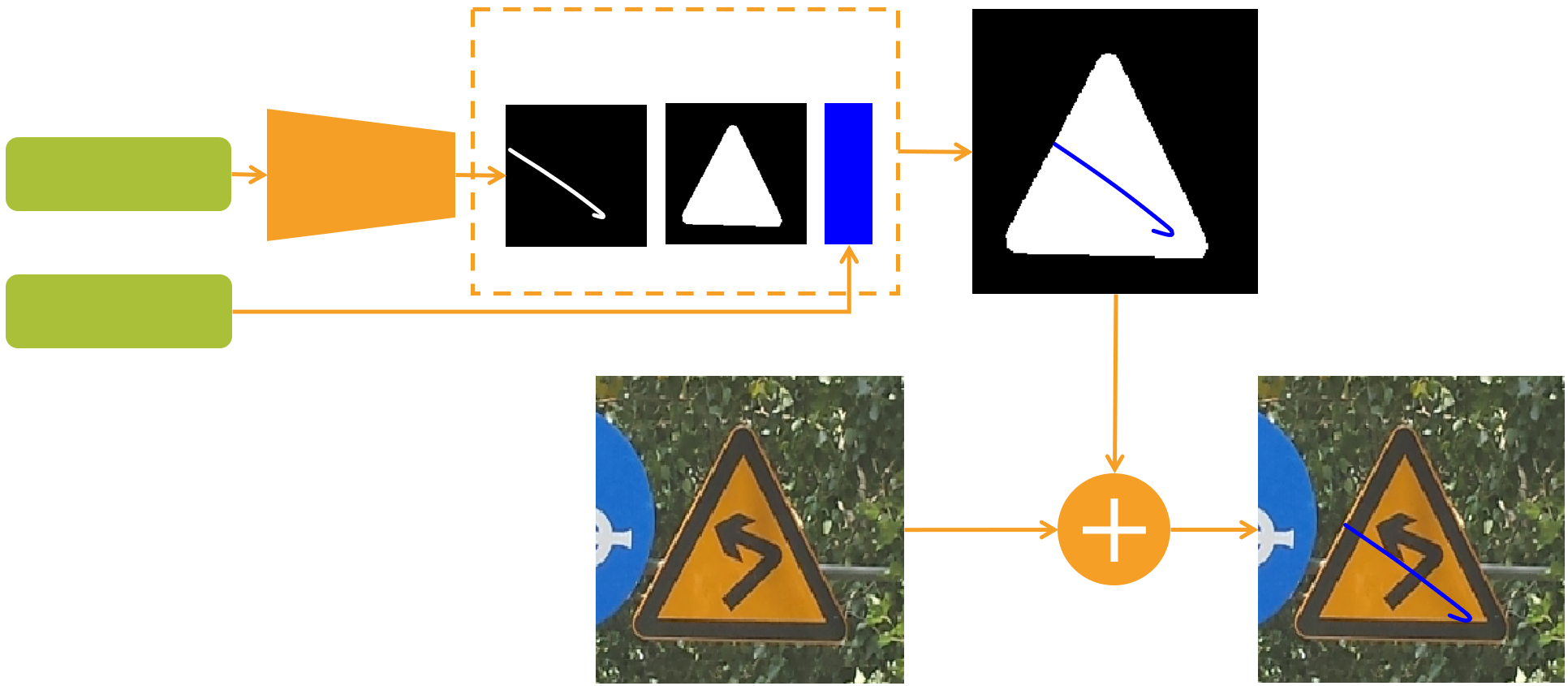}}
            \put(30,353){\Huge{$v_0, ..., v_5$}}
            \put(176,361){\color{bluegray} \huge{Perturbation}}
            \put(204,341){\color{bluegray} \huge{model}}
            \put(350,420){\Huge{$B$}}
            \put(460,420){\Huge{$\mathbf{r}$}}
            \put(536,420){\Huge{$c$}}
            \put(30,266){\Huge{$v_6, v_7, v_8$}}
            \put(469,0){\Huge{$\mathbf{x}$}}
            \put(885,0){\Huge{$\mathbf{x}^v$}}
        \end{picture}
    }
    
    \caption{Attack application methodology. From the parametric description of the attack, a perturbation can be constructed, masked, and then applied to the image, thus obtaining the adversarial image.}
    \label{fig:attackmodel}
\end{figure*}
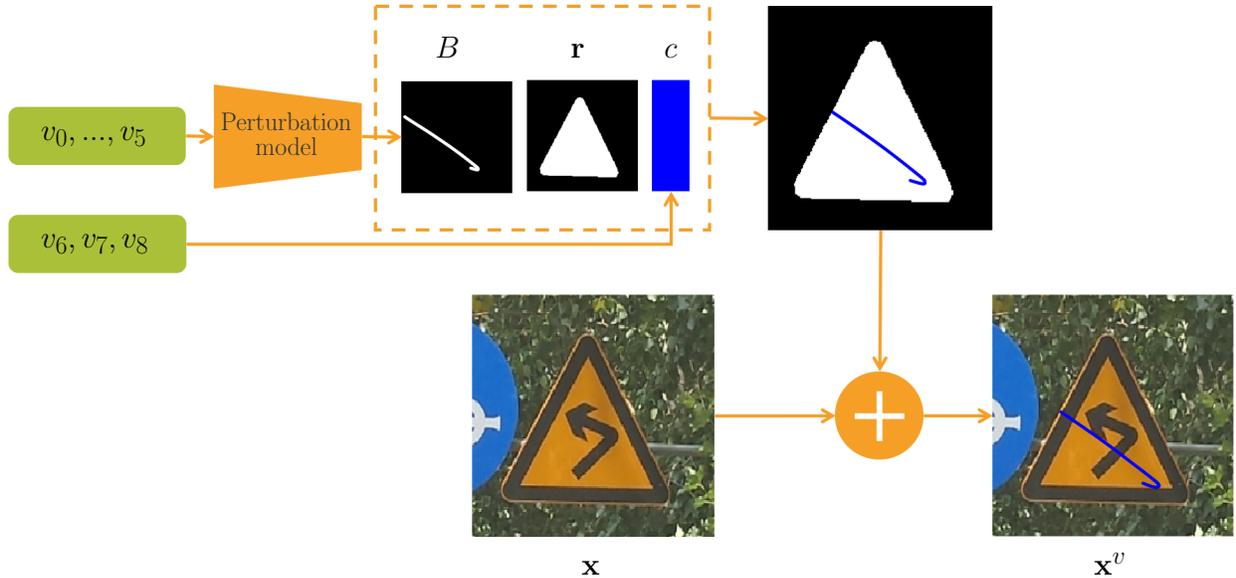

\label{sec:proposedsolution}
\begin{figure}[t!]
    \centering
    \resizebox{0.42\textwidth}{!}{
        \begin{picture}(500,159)
            \fontsize{100}{120}
            \selectfont
            \put(0,0){\includegraphics[width=500pt]{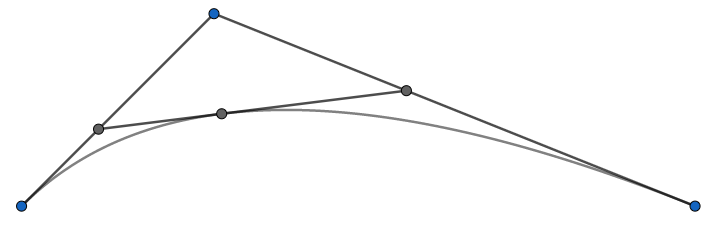}}
            \put(30,10){\Huge{$P_0$}}
            \put(155,160){\Huge{$P_1$}}
            \put(500,10){\Huge{$P_2$}}
            \put(160,50){\Huge{$B(0.4)$}}
        \end{picture}
    }
    
    \caption{Structure and construction scheme of a second-order Bézier curve $B(t)$ from its control points $P_0,P_1,P_2$. The curve is defined by interpolation of the three points by using a control parameter $t \in [0,1]$. In the image, we depict an example location of the curve $B(0.4)$.}
    \label{fig:bezier}
\end{figure}
The model we adopt to describe Adversarial Scratches is Bézier curves~\cite{hazewinkel2011encyclopaedia}. Bézier curves are polynomial segments, as such, they are continuous and continuously differentiable. While our solution is general and accounts for Bézier of any order, we illustrate our method for second-order curves, which have shown to perform best in our experimental evaluation. A second-order Bézier~(Figure~\ref{fig:bezier}) is defined as:
\begin{equation}
\label{eq:bezier}
B(t) = (1-t)^2P_0 + 2(1-t)tP_1 + t^2P_2 ,\, 0 \leq t \leq 1 .
\end{equation}

As shown in Figure~\ref{fig:bezier}, second-order Bézier curves are solely defined by their control points $P_0, P_1, P_2$, such that they start in $P_0$ and end in $P_2$, with point $P_1$ regulating their path. Thus, each scratch is identified by a vector $v \in \mathbb{R}^{9}$ (Table \ref{tab:parametrization}), where six parameters indicate the location of the Bézier's control points expressed in image coordinates, and three parameters indicate the $R, G, B$ color components of the scratch. Following common practice, to reduce the size of the search space, we only consider extreme intensity values $\{0, 255\}$, meaning that only $8$ $RGB$ triplets are available.

We chose Bézier curves because, despite their compact parametric representation, they can express a wide range of shapes. The intuition behind their adversarial nature is that these curves introduce patterns in the image to which CNNs are sensitive.
Crucially, Bézier curves possess the property of being arbitrarily subdividable (see Section \ref{sec:scratchclipping}). This allows to clip the perturbation to the target region, as shown in Figure \ref{fig:attackmodel}, and to bound them in $L_0$ terms.
These properties, in turn, allow deployability of Adversarial Scratches.

Our experiments confirm that Adversarial Scratches are a powerful attack against CNN classifiers. Moreover, our perturbations can in principle be applied on a target by using a marker or a spray can, as opposed to patches which are only applicable through a printed sticker. 
Lastly, we note that while we have defined perturbations consisting of a single scratch, our formulation is more general. When more than one scratch is to be applied, the parameter vector $v$ is expanded so that $v \in \mathbb{R}^{9\,n}$. In this case, the perturbation will be in the form of $n$ Bézier curves, as discussed in Section \ref{sec:morescratches}.

\subsection{Search Strategy}
\label{sec:optimizers}
The optimization problem described in Section \ref{sec:proposedsolution} is solved by finding a parameter configuration $v$ that leads to a successful adversarial attack. However, in the black-box attack scenario, the gradient of $\mathcal{L}_f$ is unknown. Therefore, search can only be performed by using gradient-free optimizers. Aside from being limited to this category of optimizers, the process is entirely transparent to the particular optimizer chosen to compute the perturbation.

We tested four popular gradient-free optimizers: random search (RS)~\cite{zabinsky2009random}, particle swarm optimization (PSO)~\cite{zambrano2013SPSO}, differential evolution (DE)~\cite{Price2013}, and neuro-genetic optimization (NGO)~\cite{nevergrad}. RS and DE were chosen as they were used in related works~\cite{croce2020sparse, ghosh2022black}. PSO and NGO were chosen as they represent evolutions of genetic algorithms such as DE. We discuss our choice of optimizer in Section~\ref{sec:exp_parametrization}.

\subsection{Attack Procedure}
Algorithm~\ref{algorithm:procedure} describes how the attack is performed on an image sample. For clarity, we only describe the application of a single scratch. Section~\ref{sec:morescratches} details the extension of this procedure to multiple scratches.
The required inputs are: the image $\mathbf{x}$ which we want to perturb, its target class $y$, the query threshold $MAX\_ ITER$, the model $f$ which we want to attack, the $optimizer$ that will solve the optimization problem (\ref{eq:margin_loss}), the $L_0$ bound $k$, and the region $\mathbf{r}$ used to restrict the attack only to the pixels belonging to the target object.

\begin{algorithm}[t!]
    \begin{algorithmic}[1]
        \Require $\mathbf{x}$, $y$, $MAX\_ITER$, $f$, $optimizer$, $k$, $\mathbf{r}$
        \State $iter \leftarrow 0$
        \State $C \leftarrow \argmax_{i} f(\mathbf{x})_i$ \Comment{Predict class for original image}
        \For{$i=1$ to $MAX\_ITER$}
            \State $v \leftarrow optimizer.ask()$ \Comment{Obtain a solution from the optimizer}
            \State $v^* \leftarrow clip(v, k, \mathbf{r})$ \Comment{Clip the scratch (Algorithm \ref{algorithm:filtering})}
            \State $B^* \leftarrow \text{Bézier}(v^*_0, ..., v^*_5)$ \Comment{Sample Bézier}
            \State $c \leftarrow (v^*_6, v^*_7, v^*_8)$ 
            \LeftColComment{~~Apply scratch to obtain adversarial image $x^{v^*}$}
            \State $\mathbf{x}^{v^*} \leftarrow \mathbf{x}$
            \For{$j=1$ to $|B^*|$}
                \State $\mathbf{x}^{v^*}[b^*_j] \leftarrow c$ \Comment{Set pixels on the scratch support to the scratch color}
            \EndFor
            \State $l \leftarrow f_y(\mathbf{x}^{v^*}) - \underset{z \neq y}{\mathrm{max}}\, f_z(\mathbf{x}^{v^*})$ \Comment{Feed $\mathbf{x}^{v^*}$ to the model and compute loss}
            \If{$l < 0$}
                \State $iter \leftarrow i$
                \State $\textbf{break}$
            \EndIf
            \State $optimizer.tell(v, l)$ \Comment{Update the optimizer information}
        \EndFor
        \If{$iter > 0$}
            \State $\textbf{successful attack in [iter] iterations}$
        \Else
            \State $\textbf{unsuccessful attack}$
        \EndIf
    \end{algorithmic}
    \caption{general attack procedure}
    \label{algorithm:procedure}
\end{algorithm}

In the first phase, the iteration count is initialized and the model's prediction for the original sample is computed (Lines \texttt{1-2}). 
Then, the algorithm proceeds with an iterative procedure that loops until either a valid adversarial sample is found or when a query limit is reached (Lines \texttt{3-18}). Within this loop, the first step is to request a solution from the optimizer through the standard ask-and-tell interface. By calling \emph{optimizer.ask} (Line \texttt{4}), the optimizer produces a candidate solution in the form of a vector $v$, which is detailed in Table~\ref{tab:parametrization} for second-order Bézier curves. 
Because the optimizer is not constrained to satisfy (\ref{equation:l0_constraint}) and (\ref{equation:localization_constraint}), $v$ could represent a scratch which doesn't meet the $L_0$ bound or the localization constraints given by the region $\mathbf{r}$.
To satisfy the above constraints, the $clip$ operation (Line \texttt{5}) returns a second parameter configuration $v^*$ that corresponds to a scratch which is a subset of the original one. This is possible since the set of Bézier curves is closed with respect to arbitrary subdivision~\cite{barsky1985arbitrary}. The details of the clipping and parameter update procedure are described in Algorithm \ref{algorithm:filtering}.
Given $v^*$, we compute \emph{i)} the support set of the scratch $B^* = \{(x_1, y_1), \dots\}$ by sampling curve (\ref{eq:bezier}) and \emph{ii)} its color $c = \{v^*_6, v^*_7, v^*_8\}$ (Lines \texttt{6-7}).
The tentative adversarial image $\mathbf{x}^{v^*}$ is then obtained by changing the color of $\mathbf{x}$ for each pixel location in the scratch support $B^*$ to the scratch color $c$ (Lines \texttt{8-11}):
\begin{equation}
\label{eq:singlescratch}
\mathbf{x}^{v^*}[i,j]=
\begin{cases}
    c      & \text{if } (i,j) \in B^* \\
    \mathbf{x}[i,j] & \text{otherwise}
\end{cases}
.
\end{equation}

Thus, we compute the margin loss~(\ref{eq:margin_loss}) by feeding the tentative adversarial image $\mathbf{x}^{v^*}$ to the target model $f$ (Line~\texttt{12}), and if the attack is successful (\ref{eq:margin_loss_target}), the procedure ends (Line~\texttt{20}). Otherwise, the state of the optimizer is updated (Line~\texttt{17}) by providing the loss value for the given configuration, and the procedure continues to the next iteration. If no adversarial sample can be computed within the query limit, the attack is unsuccessful and the algorithm terminates (Line~\texttt{22}). In our experiments, in line with~\cite{croce2020sparse}, the query limit was set to $10\,000$. This is better discussed in Section~\ref{sec:experimental_setup}.

We note that the clipping operation transforms the parameter vector $v$ before the sample is perturbed. Thus, when updating the state of the optimizer, the parameter configuration $v$ is associated to the margin loss $\mathcal{L}_f(\mathbf{x}, \mathbf{x}^{v^*})$ where the argument is the sample with the clipped perturbation $\mathbf{x}^{v^*}$ instead of $\mathbf{x}^{v}$.
This choice is justified since clipping does not modify feasible solutions, therefore, for all feasible $\mathbf{x}^{v}$, $\mathcal{L}_f(\mathbf{x}, \mathbf{x}^{v}) = \mathcal{L}_f(\mathbf{x}, \mathbf{x}^{v^*})$.

\section{Implementation Details}
\label{sec:implementation}
In this section, we provide practical details regarding the generation of adversarial examples as shown in Algorithm~\ref{algorithm:procedure}, and discuss the extension of the attack to multiple scratches.

\subsection{Scratch Clipping}
\label{sec:scratchclipping}
Since the optimizer is agnostic to the underlying perturbation model, it may propose parameters $v$ that may represent scratches that do not satisfy the deployability constraints. We address this issue by clipping the scratch to a connected segment of the original one, such that the clipped scratch satisfies the $L_0$ constraint and is entirely contained in the target region.
The clipping procedure, described in Algorithm \ref{algorithm:filtering}, receives as input the configuration $v$, the $L_0$ bound $k$, and the region $\mathbf{r}$, and returns a new parameter configuration $v^*$. 
This procedure modifies the coordinates of the control points of the Bézier, leaving its color unchanged. For second-order Bézier curves (Table \ref{tab:parametrization}), this means modifying parameters $v_0, \dots, v_5$.

\begin{algorithm}[t!]
    \caption{Scratch clipping}
    \label{algorithm:filtering}
    \begin{algorithmic}[1]
        \Require $v$, $k$, $\mathbf{r}$
        \State $P_0 = \{v_0, v_1\}, P_1 = \{v_2, v_3\}, P_2 = \{v_4, v_5\}$ \Comment{Input Bézier control points}
        \State $flag \leftarrow 0, d \leftarrow 0, B^* \leftarrow \emptyset$ \Comment{Initialize flags and scratch support}
        \LeftColComment{Sample the Bézier to obtain points and their parametric position}
        \State $B, T \leftarrow \text{Bézier}(v_0, ..., v_5)$
        \For{$j=1$ to $|B|$}
            \If{$flag == 0$ \texttt{and} $\mathbf{r}[B]==1$} \Comment{Beginning of the segment found}
                \State $p \leftarrow j$
                \State $flag \leftarrow 1$
            \EndIf
            \If{$flag == 1$}
                \If{$\mathbf{r}[B]==0$ \texttt{or}  $d==k$} \Comment{End of the segment is reached}
                    \State $q \leftarrow d-1$
                    \State $\textbf{break}$
                \Else
                    \State $d$\texttt{++}
                    \State $B^* \leftarrow B^* \bigcup b_j$ \Comment{Add the location to the support}
                \EndIf
            \EndIf
        \EndFor
        \LeftColComment{Split Bézier and return new parameter configuration}
        \vspace{1pt}
        \State $P^*_0 \leftarrow (b^*_{1,x}, b^*_{1,y})$
        \vspace{1pt}
        \State $P^{'}_1 \leftarrow (1-t_p) P_1+t_p P_2$
        \vspace{1pt}
        \State $P^{''}_1 \leftarrow (1-t_{p+q}) b_p+t_{p+q} P^{'}_1$
        \vspace{1pt}
        \State $P^*_2 \leftarrow (b^*_{|B^*|,x}, b^*_{|B^*|,y})$
        \vspace{1pt}
        \State $v \leftarrow \{P^*_{0,x}, P^*_{0,y}, P^{''}_{1,x}, P^{''}_{1,y}, P^*_{2,x}, P^*_{2,y}, v_6, v_7, v_8\}$
        \vspace{1pt}
        \State \Return $v^*$
    \end{algorithmic}
\end{algorithm}

In practice, we sample the scratch identified by $v$ and obtain its support $B$ (Lines \texttt{1-3}). Then (Lines \texttt{4-18}), we select the longest contiguous subset $B^* = \{b_p, b_{p+1}, \dots, b_{p+q}\}$ of $B$ 
which is entirely contained in the region $\mathbf{r}$, starting at the first pixel\footnote{The ordering is given by the parametric location $t$ along the Bézier curve (\ref{eq:bezier})} of $B$ belonging to the mask.
The set of all Bézier curves is closed with respect to arbitrary subdivision \cite{barsky1985arbitrary}, thus, given any two points $P^*_0$, $P^*_2$  along a Bézier $B$, it is possible to compute the parameters of a new Bézier $B^*$ which is a segment of $B$ starting in $P^*_0$ and ending in $P^*_2$.
We achieve this by using the arbitrary subdivision procedure \cite{barsky1985arbitrary} up to two times (Lines \texttt{19-20}). By the first subdivision, we remove the segment $\{b_1, ..., b_p\}$, obtaining a Bézier starting in $P^*_{0}$, ending in $P_2$, and having control point $P^{'}_1$. By the second subdivision we remove $\{b_{p+q}+1, ..., b_{|B|}\}$, finally obtaining a Bézier starting in $P^*_{0}$, ending in $P^*_{1}$, having control point $P^{''}_1$.
Lastly (Lines \texttt{21-22}), we return the new parameter configuration $v^*$ representing the clipped scratch. By construction, the new Bézier satisfies the $L_0$ and localization constraints.

\subsection{Multiple Scratches}
\label{sec:morescratches}
Algorithm~\ref{algorithm:procedure} describes how to apply a single Adversarial Scratch. We may however want to perturb $\mathbf{x}$ by using $n$ scratches. For second-order Bézier curves, the parameter vector describing $n$ scratches will be $v \in \mathbb{R}^{9\,n}$, where each tuple of $9$ parameters represents one scratch, having support $B_s$ and color $c_s, s = 1...n$.
Before application, each scratch needs to be clipped according to Algorithm~\ref{algorithm:filtering} to obtain the clipped supports $B^*_s$ and parameter configuration $v^*$, then, the scratches are applied in sequence, possibly overwriting already perturbed pixels when these overlap. We extend (\ref{eq:singlescratch}) for the case with $n>1$ scratches, so that the image where scratches $1 \dots s$ are applied is defined as follows:
\begin{equation}
\mathbf{x}^{(s)}[i,j]=
\begin{cases}
    c_s      & \text{if } (i,j) \in B^*_s \\
    \mathbf{x}^{(s-1)}[i,j] & \text{otherwise}
\end{cases}
,
\end{equation}
where $\mathbf{x}^{(0)} = \mathbf{x}$ and the end result is $\mathbf{x}^{v^*} = \mathbf{x}^{(n)}$.\

Increasing the number of scratches makes the attack more powerful, as it allows the perturbation to cover different regions within $\mathbf{r}$, but comes at the cost of deployability, as one would need to account for the relative positions between scratches.
In practice, we limit to a maximum of $n=5$ Adversarial Scratches, which we deem to be a reasonable bound allowing for powerful yet deployable attacks. In Section \ref{sec:exp_parametrization} we study the effects of using attacks with varying number of scratches.

\section{Experiments}
In this section, we analyze the performance of Adversarial Scratches in a variety of experiments, utilizing the ImageNet \cite{russakovsky2015imagenet} and TSRD \cite{TSRD} datasets. First, we describe the employed experimental setup (Section~\ref{sec:experimental_setup}), then, we compare Adversarial Scratches to state-of-the-art deployable attacks, presenting experiments on the ImageNet and TSRD datasets, and against Microsoft Cognitive Services API (Section~\ref{sec:experiments_deployable}). We then compare Adversarial Scratches to non-deployable state-of-the-art attacks (Section~\ref{sec:experiments_non_deployable}), as they represent an ideal performance reference for $L_0$ attacks.
Lastly, we present a thorough exploration of several possible configurations of Adversarial Scratches (Section~\ref{sec:exp_parametrization}).
All experiments target ResNet-50 classifiers, except where otherwise stated. Since many considered attacks have a stochastic component, we average results from five runs with different random seeds, also reporting standard deviations across runs.

\subsection{Experimental Setup}
\label{sec:experimental_setup}
We introduce our experimental evaluation framework, and detail the figures of merit used throughout the experiments.

\noindent\textbf{Framework:} 
We have developed a flexible Python framework to run our experiments. The framework allows to design, implement, and execute adversarial attacks on images. The model to be attacked, the optimizer, and the perturbation model are modular, and can be combined to perform a variety of tests in different combinations. The modules communicate via straightforward interfaces, allowing seamless integration and extension to many attack methodologies, including those implemented in Sparse-RS \cite{croce2020sparse}. We release our code to the public as a platform to test adversarial attacks on neural networks. 

\noindent\textbf{Metrics:}
To measure the performance of the attacks, we compute fooling rate (FR), average queries (AQ), and median queries (MQ). These metrics are significantly influenced by the query limit, thus, following common practice, all methods are compared using the same query limit of $10\,000$. The metrics are defined as follows:
\begin{itemize}[leftmargin=*]
    \item{\textbf{Fooling Rate (FR)}}
    is the fraction of image samples for which the attack was successful within the query limit, out of all samples that were subject to the attack. Higher FR indicates that the attack was more successful. \
    
    \item{\textbf{Average Queries (AQ)}}
    is the average number of queries needed to craft a successful perturbation. Lower AQ values indicate that the attack needs fewer attempts, on average, to find a perturbation that deceives the model.\footnote{Both AQ and MQ are computed over images where the attack was successful within the query limit.\label{fn:metrics}}
    
    \item{\textbf{Median Queries (MQ)}}
    is the median number of queries needed to craft a successful perturbation.
    MQ is useful since query requirements may greatly vary between samples. Therefore, AQ and MQ, in conjunction, allow to better understand the distribution of query requirements.\footref{fn:metrics}
    
\end{itemize}

\noindent\textbf{Considered Methods:}
In our experiments, we compare Adversarial Scratches to several state-of-the-art $L_0$ attacks. Following \cite{croce2019sparse} and \cite{modas2019sparsefool}, we do not consider $L_1$, $L_2$, and $L_\infty$ attacks, since given a specific $L_0$ bound it is impossible to define $L_1$, $L_2$, and $L_\infty$ bounds that would result in comparable perturbations. This is further justified by our interest in deployable attacks. Indeed, as stated in Section~\ref{sec:intro}, $L_1$, $L_2$, and $L_\infty$ attacks cannot be deployed.

\noindent\textbf{Optimizer:}
Results for Adversarial Scratches have been obtained using the NGO optimizer, as it shows the best performance amongst the considered ones. In Section~\ref{sec:exp_parametrization} we compare the performance of various optimizers.

\subsection{Comparison Against Deployable Attacks}
\label{sec:experiments_deployable}
We analyze the performance of Adversarial Scratches against state-of-the-art deployable attacks on the widely used ImageNet dataset. To further investigate the deployability of our attacks, we perform experiments on the TSRD traffic sign dataset and we deploy our attack against the publicly available Microsoft Cognitive Services API.

\begin{figure*}[t]
    \centering
    \renewcommand\tabcolsep{4pt}
    \resizebox{0.8\textwidth}{!}{
        \begin{tabularx}{\textwidth}{p{8pt}p{0.1pt}YYYp{25pt}}
            ~ & & \small Patch & \small Adversarial Scratches & \small ``any-pixel'' & \\ 
            \rotatebox[origin=c]{90}{\textbf{$L_0=400$}} &
            \multicolumn{4}{l}{
                \begin{tabular}{YYY}
                    \setlength{\fboxsep}{0pt}\setlength{\fboxrule}{1pt}\fcolorbox{goodgreen}{black}{\includegraphics[width=0.275\textwidth,valign=T]{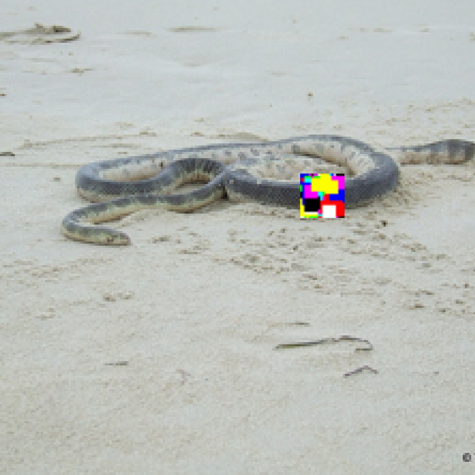}} & 
                    \setlength{\fboxsep}{0pt}\setlength{\fboxrule}{1pt}\fcolorbox{goodgreen}{black}{\includegraphics[width=0.275\textwidth,valign=T]{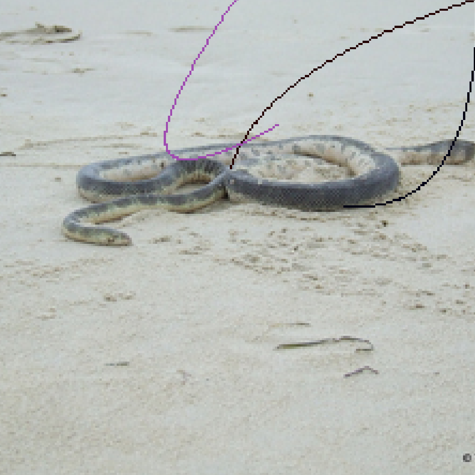}} & 
                    \setlength{\fboxsep}{0pt}\setlength{\fboxrule}{1pt}\fcolorbox{badred}{black}{\includegraphics[width=0.275\textwidth,valign=T]{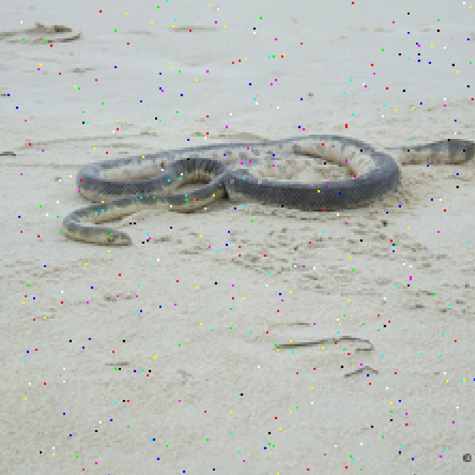}} \vspace{2pt} \\
                \end{tabular}
            }
            & 
            \begin{tabular}{l}
                 \color{goodgreen} $\square$ \small Deployable \\
                 \color{badred} $\square$ \small Non-deployable \\
            \end{tabular}
            \\
            ~ & & \noindent \rightskip=0pt \justifying \scriptsize The attack perturbs a square region of size $\sqrt{k}$ pixels& 
            \noindent \rightskip=0pt \justifying \scriptsize The attack places three scratches, one pixel wide and $\frac{k}{3}$ pixels long& 
            \noindent \rightskip=0pt \justifying \scriptsize The attack modifies $k$ pixels anywhere on the image & \\ 
        \end{tabularx}
    }
    \vspace{-5pt}
    \caption{Different attack structures with the same $L_0$ constraint of 400 perturbed pixels. On the left, a patch attack \cite{croce2020sparse, rao2020adversarial, yang2020patchattack}, in the middle, our 3-scratch attack, and on the right, the ``any-pixel'' attack \cite{croce2020sparse}.}
  \label{fig:results400}
\end{figure*}

\noindent\textbf{Considered Methods:}
We compare against Patch-RS~\cite{croce2020sparse}, PatchAttack~\cite{yang2020patchattack}, and LOAP w/ GE~\cite{rao2020adversarial, croce2020sparse}, as these are the best performing deployable attacks in the literature. All of these attacks are based on a perturbation model that modifies a square patch (Figure \ref{fig:results400}, left), differing only in how the patch is defined and optimized. Patch-RS uses random search to overlap colored rectangles within the patch area, while PatchAttack uses an RL agent to optimize a textured adversarial patch. Lastly, LOAP is originally a white-box method that optimizes the adversarial patch through gradient information, which is adapted to the black-box scenario by using Gradient Estimation, as in~\cite{croce2020sparse}, resulting in the black-box LOAP w/ GE.

\noindent\textbf{Experiments on ImageNet:}
This experiment provides a comparison to the state-of-the-art in a standard setting with no restrictions on the target region. 

We compare $20 \times 20$ patches generated with Patch-RS, LOAP w/ GE, and PatchAttack (total $L_0=400$) to an attack composed of three $L_0 = 133$ scratches. $L_0$ bounds are thus comparable, as this Adversarial Scratches attack is constrained to a total $L_0=399$ (Figure~\ref{fig:results400}, center). Since our experimental setup is equivalent to that of Sparse-RS~\cite{croce2020sparse}, results for Patch-RS, LOAP w/ GE, and PatchAttack are taken as reported from their paper.

\begin{table*}[h]
    \centering
    \caption{Comparison to deployable attacks. Results for Patch-RS, LOAP w/ GE and PatchAttack are reported from~\cite{croce2020sparse}, which uses the same experimental setup. Results are computed using a query limit of $10\,000$.}
    \label{tab:structured_400}
    \resizebox{1.0\textwidth}{!}{
        \begin{tabular}{|l|c|c|c|l|}
            \hline
            Attack & FR & AQ & MQ & Perturbation model\\
            \hline
            LOAP w/ GE & 40.6\% $\pm$0.1\% & 6870 $\pm$10 & 10000 $\pm$0 & 20$\times$20 patch\\
            PatchAttack & 49.6\% $\pm$1.2\% & 5722 $\pm$64 & 5280 $\pm$593 & 20$\times$20 patch\\
            Patch-RS & 79.5\% $\pm$1.4\% & 2808 $\pm$89 & 438 $\pm$68 & 20$\times$20 patch\\
            Adversarial Scratches & \textbf{97.9\% $\pm$0.3\%} & \textbf{302 $\pm$38} & \textbf{27 $\pm$3} & Three 133px long Bézier\\
            \hline
        \end{tabular}
    }
\end{table*}

Table~\ref{tab:structured_400} shows that Adversarial Scratches have much higher fooling rate while requiring significantly fewer queries than other compared methods. Moreover, we note that Adversarial Scratches are typically shorter than their nominal length. Indeed, in this experiment, Adversarial Scratches modify on average $331.9$ pixels, fewer than the allowed $399$ pixels. Other attacks, in contrast, always modify all $400$ allowed pixels. These results show that Adversarial Scratches are better than current state-of-the-art deployable attacks in all considered metrics.

\noindent\textbf{Experiments on TSRD:}
To better tests the deployability of the attacks, we design an experiment on the TSRD~\cite{TSRD} dataset where attacks are only allowed to perturb pixels in a target region which corresponds to a traffic sign.
The target model is a ResNet-50 classifier finetuned to $98\%$ test accuracy on the TSRD dataset.

\begin{figure}[t!]
    \centering
    \renewcommand\tabcolsep{4pt}
    \resizebox{0.48\textwidth}{!}{
        \begin{tabularx}{\textwidth}{p{8pt}YYY}
            ~ & R-Patch-RS & Adversarial Scratches & Target region\\ 
            \multirow{9}[6]{*}{\rotatebox[origin=c]{90}{\textbf{$L_0=400$}}} & \setlength{\fboxsep}{0pt}\setlength{\fboxrule}{0.2pt}\fbox{\includegraphics[width=0.30\textwidth,valign=T]{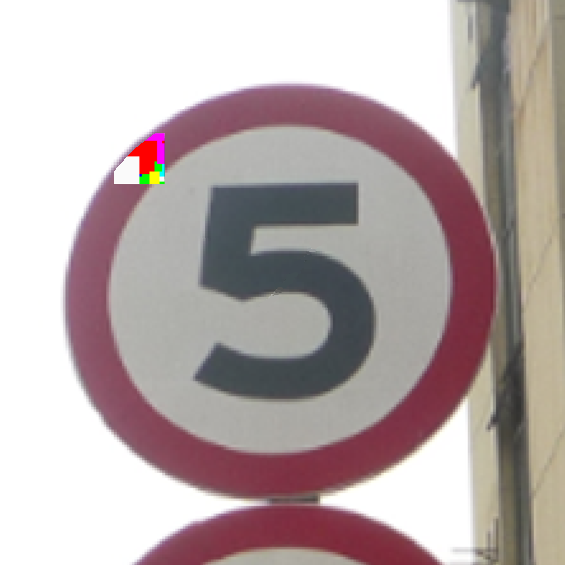}} & 
            \setlength{\fboxsep}{0pt}\setlength{\fboxrule}{0.2pt}\fbox{\includegraphics[width=0.30\textwidth,valign=T]{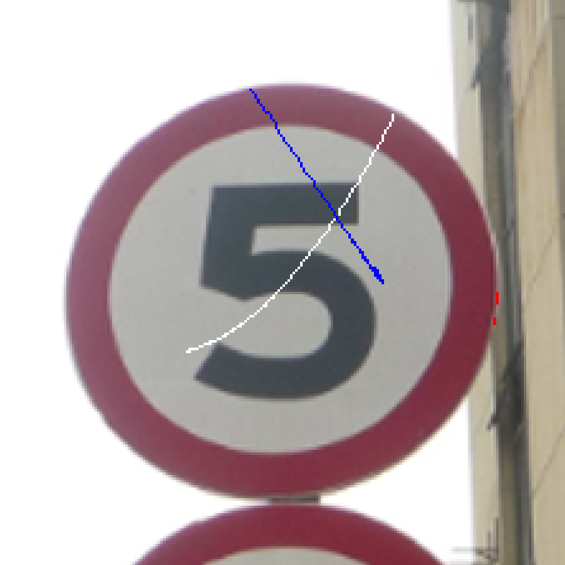}} & 
            \setlength{\fboxsep}{0pt}\setlength{\fboxrule}{0.2pt}\fbox{\includegraphics[width=0.30\textwidth,valign=T]{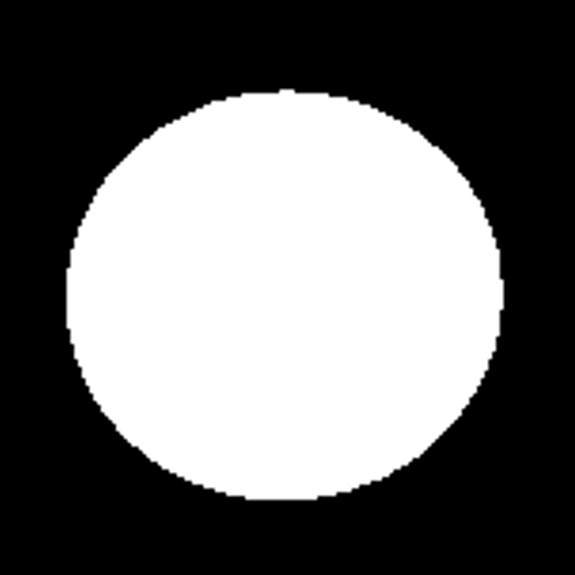}}\\
            
        \end{tabularx}
    }

    \caption{Deployable attacks applied to the TSRD street sign dataset. Left: a $20 \times 20$ R-Patch-RS attack. Centre: an attack composed of two Adversarial Scratches with $L_0$ bound set to $ 200$. Both attacks are constrained to the street sign's region $\mathbf{r}$, which is depicted on the right.}
    \label{fig:resultstsrd}
\end{figure}

\noindent\textbf{\textit{Considered Methods:}}
For this experiment, we only compare to Patch-RS as it was shown (Table~\ref{tab:structured_400}) to be the best performing deployable attack amongst the tested ones.
However, in its base form, Patch-RS cannot be localized within a target region. To solve this issue, we have developed a spatially localizable version of Patch-RS, namely, R-Patch-RS. This was achieved with minimal modifications to Sparse-RS' code.
Figure~\ref{fig:resultstsrd} shows an example of Adversarial Scratches~(center) and R-Patch-RS~(left) attacks on TSRD images.

\noindent\textbf{\textit{Dataset Preparation and Model Setup:}}
The TSRD \cite{TSRD} dataset is composed of $6164$ traffic sign images, divided in $58$ classes. For this experiment, we have created target regions by manually segmenting non-occluded pixels of traffic signs from more than $100$ TSRD samples. We remark that this manual annotation was only performed to conform to a realistic scenario where the attack perturbation is limited to the region described by the street sign.

Since the TSRD dataset includes augmented and duplicate images, to favor unique images, the images used for this experiment were selected manually.
Selected images and their augmented versions were never used for training, ensuring that attacks are performed on images the model has never seen.
We publicly release these segmentation masks, together with information on how to match them with images in the original TSRD dataset.

\begin{table*}[t]
    \centering
    \caption{Performance of attacks on the TSRD dataset when restricted to the target's pixels.}
    \label{tab:tsrd}
    \resizebox{1.0\textwidth}{!}{
        \begin{tabular}{|l|c|c|c|l|}
            \hline
            Attack & FR & AQ & MQ & Comment\\
            \hline
            R-Patch-RS & 97.5\% $\pm$1.4\% & 366 $\pm$58 & 195 $\pm$15 & $20\times20$ patch ($L_0 = 400$)\\
            Adversarial Scratches & \textbf{100\% $\pm$0.0\%} & \textbf{68 $\pm$31} & \textbf{6 $\pm$1} & Three Bézier each $133$px long ($L_0 = 399$)\\
            \hline
        \end{tabular}
    }
\end{table*}

\noindent\textbf{\textit{Results:}}
Table \ref{tab:tsrd} shows that, although the attacks are constrained to only affect pixels belonging to the traffic sign in the image, both Adversarial Scratches and R-Patch-RS are very successful, 
and ultimately results in almost all samples being successfully attacked.
We find that Adversarial Scratches have better AQ and MQ than R-Patch-RS. Most notably, median queries drop from 195 to 6, meaning that, by using Adversarial Scratches, we could attack half the traffic signs in the dataset with just 6 attempts.

\begin{figure*}[t!]
    \centering
    \renewcommand\tabcolsep{0pt}
    \renewcommand\arraystretch{1.2}
    \resizebox{1.0\textwidth}{!}{
        \begin{tabularx}{\textwidth}{YYYY} %
            ~ &  \vspace{0.1pt} \cellcolor{lorange}Original &  & \\ 
            \multirow{6}[6]{*}{\,\,\,\,\,\,\,\,\,\,\,\,\,Input image} & \cellcolor{lorange}\includegraphics[width=0.25\textwidth,valign=T]{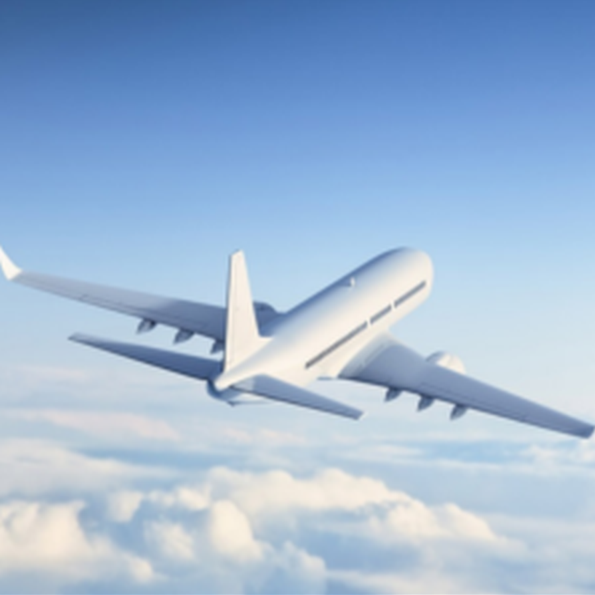} & \includegraphics[width=0.25\textwidth,valign=T]{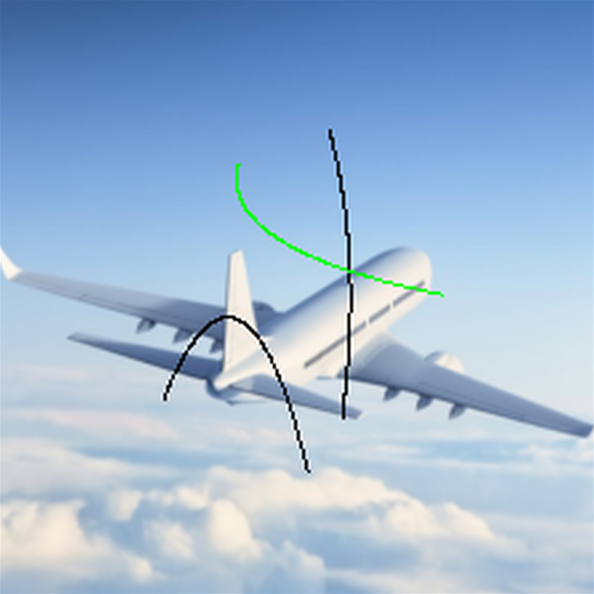} & \includegraphics[width=0.25\textwidth,valign=T]{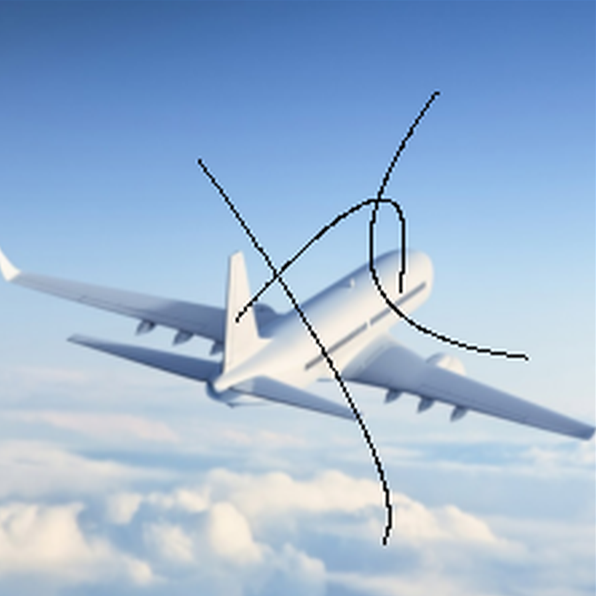}\\
            \cellcolor{gray1} Predicted caption & 
            \cellcolor{lorange} \tiny A plane flying in the sky & 
            \cellcolor{gray1} \tiny A green and white rocket & 
            \cellcolor{gray1} \tiny A wind turbine with a blue sky \\
            \,\,\,\,\,\,\,\,\,\,\,\,\,\,\,\,\cellcolor{gray2} Confidence & \cellcolor{lorange} $0.512$ & \cellcolor{gray2} $0.219$ & \cellcolor{gray2} $0.375$ \\
            \,\,\,\,\,\,\,\,\,\,\,\,\,\,\,\,\,\,\,\cellcolor{gray1} Iterations & \cellcolor{lorange} $0$ & \cellcolor{gray1} $6$ & \cellcolor{gray1} $25$ \\
            \,\,\,\,\,\,\,\,\,\,\,\,\,\,\,\,\,\,\,\,\,\,\,\,\,\,\,\,\,\,\,\,\,\,\,\,\,\,\cellcolor{gray2} $L_0$ & \cellcolor{lorange} $0$ & \cellcolor{gray2} $341$ & \cellcolor{gray2} $435$ \\
            \multirow{6}[6]{*}{\,\,\,\,\,\,\,\,\,\,\,\,\,Input image} & \includegraphics[width=0.25\textwidth,valign=T]{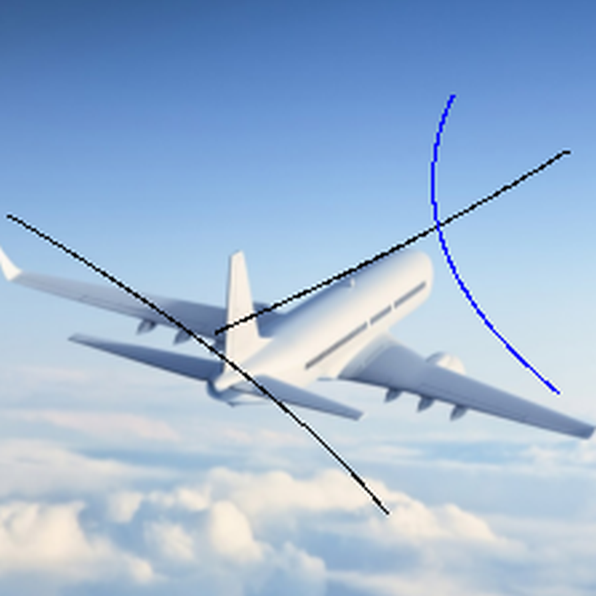} & \includegraphics[width=0.25\textwidth,valign=T]{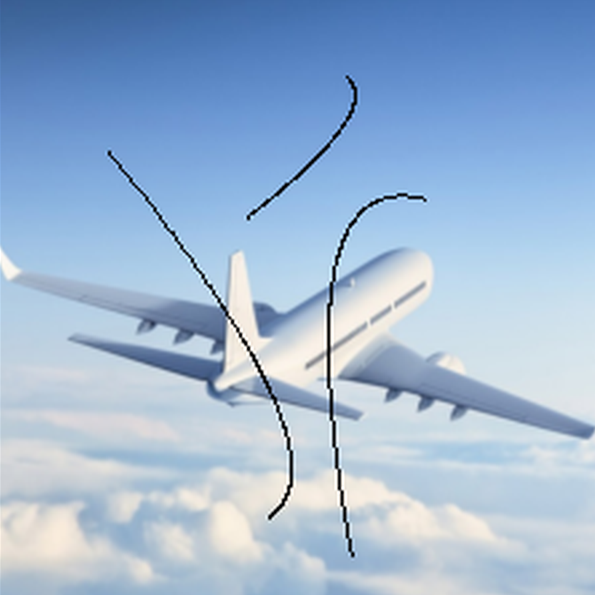} & \includegraphics[width=0.25\textwidth,valign=T]{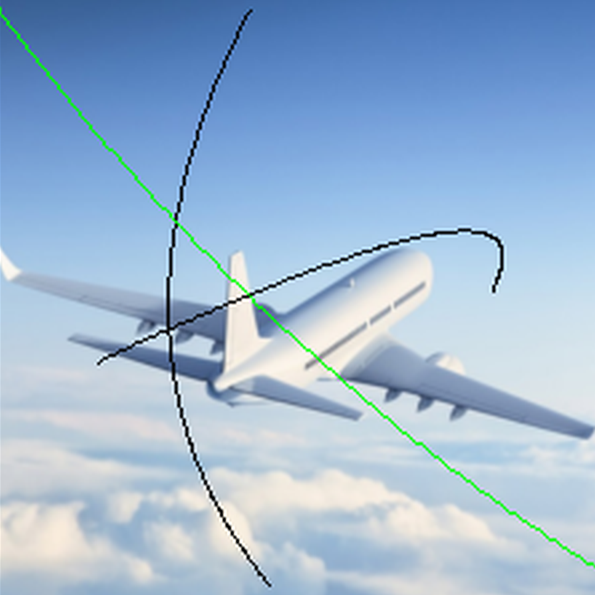}\\
            \cellcolor{gray1} Predicted caption & 
            \cellcolor{gray1} \tiny A group of thin thin thin thin thin thin thin thin thin thin thin thin thin thin thin thin & 
            \cellcolor{gray1} \tiny A close-up of a barbed wire fence & 
            \cellcolor{gray1} \tiny A close-up of a bug \\
            \,\,\,\,\,\,\,\,\,\,\,\,\,\,\,\,\cellcolor{gray2} Confidence & \cellcolor{gray2} $0.169$ & \cellcolor{gray2} $0.493$ &  \cellcolor{gray2} $0.539$ \\
            \,\,\,\,\,\,\,\,\,\,\,\,\,\,\,\,\,\,\,\cellcolor{gray1} Iterations & \cellcolor{gray1} $57$ & \cellcolor{gray1} $65$  & \cellcolor{gray1} $217$ \\
            \,\,\,\,\,\,\,\,\,\,\,\,\,\,\,\,\,\,\,\,\,\,\,\,\,\,\,\,\,\,\,\,\,\,\,\,\,\,\cellcolor{gray2} $L_0$ & \cellcolor{gray2} $481$ & \cellcolor{gray2} $419$ & \cellcolor{gray2} $683$ \\
        \end{tabularx}
    }
    \caption{Successful attacks against the Microsoft Cognitive Services Image Captioning API. The top left image denotes the source image, which is captioned correctly as `a plane flying in the sky'. All other images are examples of images that successfully deceived the captioning service. Note that, even though each of the three scratches composing the attack was bounded to $L_0 \leq 500$, the overall $L_0$ difference with respect to the original is much lower than $1500$.}
    \label{tab:microsoft-attacks}
\end{figure*}

\noindent\textbf{Experiments on Microsoft Cognitive Services API:}
We perform an attack against the Microsoft Cognitive Services Image Captioning API~\cite{vision-api} using Adversarial Scratches. We formalize this API as a model $g$ which, given an image $\mathbf{x}$, provides a caption describing the image's content and the model's confidence $h(\mathbf{x})$. Since the margin loss (\ref{eq:margin_loss}) cannot be computed in the captioning scenario, we solve the optimization problem by minimizing the loss: 
\begin{equation}
    \mathcal{L}_{g} = h(\mathbf{x}) .
\end{equation}
The rationale behind this is to minimize the confidence to induce the model to produce wrong captions.
In this experiment, we analyze the performance of Adversarial Scratches against a real-world system, however, since the target is an online service, physical deployability is irrelevant. Thus, in this setting, we use $3$ scratches each with $L_0 \leq 500$, with no restriction to a target region.
As API calls are rate limited, it is unfeasible to perform thousands of consecutive queries, thus, we have attacked a single image and provide qualitative results in Figure~\ref{tab:microsoft-attacks}.

\noindent\textbf{\textit{Results:}}
Adversarial Scratches were able to deceive the API into generating wrong captions. Most notably, we were able to significantly change the output caption in just 6 iterations (the caption was altered also in earlier iterations but not significantly, e.g. ``A plane" was substituted by ``A jet"). Examples of these generated scratches are shown in Figure~\ref{tab:microsoft-attacks}. The attacks were performed on \texttt{September 6th, 2021}. \emph{The vulnerability was reported to Microsoft}.

\boldmath
\subsection{Comparison Against Non-deployable Attacks}
\unboldmath
\label{sec:experiments_non_deployable}
We compare the performance of Adversarial Scratches to that of other non-deployable $L_0$ attacks. Importantly, \emph{this comparison is unfair to Adversarial Scratches}, since the non-deployable attacks we compare to are more general than Adversarial Scratches and can exploit a much wider attack surface.

\noindent\textbf{Considered Methods:}
We compare to the ``any-pixel'' attack from Sparse-RS (Figure~\ref{fig:results400}, right), which uses random search to optimize a perturbation of any $k$ pixels in the image, and SimBA~\cite{guo2019simple}, which finds orthonormal directions to iteratively improve the perturbation. We chose SimBA and Sparse-RS's ``any-pixel'' attack as they are state-of-the-art attacks in the black-box, $L_0$ scenario. 

\noindent\textbf{Experimental Setup:}
Since we are comparing to non-deployable attacks, we only focus on the ImageNet dataset, with no restrictions on the target region.
We test under $L_0$ constraints of $400$ and $50$ pixels. In the $L_0=400$ scenario, we use an attack composed of three $L_0 = 133$ scratches. For the $L_0=50$ scenario, we use a single $L_0 = 50$ scratch, as shorter segments would not display features typical of scratches.
All attacks are run with query limit set to $10\,000$, exception made for SimBA, as this attack perturbs one color channel of one pixel each iteration. To enable a fair comparison, SimBA was limited to a number of iterations equal to three times the $L_0$ bound. Although this results in a lower query limit than $10\,000$, this goes in favour of SimBA as the attack can potentially modify three times more pixels.

\noindent\textbf{Results:}
Table~\ref{tab:unstructured} shows that, in the $L_0 = 400$ scenario, the fooling rate of Adversarial Scratches is comparable to the state-of-the-art, with Sparse-RS' ``any-pixel'' attack being marginally better, and SimBA being worse than both. 
The outcome is different in the very strict $L_0 = 50$ case, where the non deployable Sparse-RS is still able to achieve $83.9\%$ FR, while our attack's performance drops significantly to $55.8\%$. The performance of SimBA is also greatly reduced, demonstrating that this is an extremely constrained test scenario.
These results show that, when Adversarial Scratches can assume their typical shapes, as is the case for the $L_0 = 400$ scenario, our attack achieves surprisingly good performance, comparable even to those of non-deployable state-of-the-art attacks. This is not true for the $L_0 = 50$ scenario, where scratches are too short to display adversarial features.

\begin{table*}[t!]
    \centering
    \caption{Comparison to non-deployable $L_0$ attacks. SimBA is only run once as the provided code does not allow changing seed.}
    \renewcommand\arraystretch{1.2}
    \renewcommand\tabcolsep{4pt}
    \resizebox{1.0\textwidth}{!}{
        \begin{tabular}{|p{0pt}p{0pt}|p{0pt}p{0pt}|p{100pt}|>{\centering\arraybackslash}p{66pt}|>{\centering\arraybackslash}p{50pt}|>{\centering\arraybackslash}p{50pt}|p{165pt}|}
            \cline{5-9}
            \multicolumn{4}{l|}{} & Attack & FR & AQ & MQ & Perturbation model \\
            \cline{1-9}
            \multirow{4}[4]{*}{\rotatebox[origin=c]{90}{\textbf{$L_0$}~~~~~}} & & \multirow{2}[2]{*}{\rotatebox[origin=c]{90}{\textbf{$400$}~~}} & & SimBA & 71.3\% & 500.15 & 457 & Any 1200 pixel channels in the image\\
             & & & & Adversarial Scratches & 97.9\% $\pm$0.3\% & 302 $\pm$38 & 27 $\pm$3 & Three Bézier, each 133px long\\
             & & & & ``any-pixel'' & \textbf{99.9\% $\pm$0.2\%} & \textbf{154 $\pm$6} & \textbf{25 $\pm$1} & Any 400 pixels in the image\\
            \cline{3-9}
             & & \multirow{2}[2]{*}{\rotatebox[origin=c]{90}{\textbf{$50$}~~}} & & SimBA & 12.1\% & 72.6 & 68 & Any 150 pixel channels in the image\\
             & & & & Adversarial Scratches & 55.8\% $\pm$0.4\% & \textbf{866 $\pm$35} & \textbf{75 $\pm$7} & One Bézier, 50px long\\
             & & & & ``any-pixel'' & \textbf{83.9\% $\pm$0.8\%} & 1899 $\pm$48 & 906 $\pm$23 & Any 400 pixels in the image\\
            \cline{1-9}
        \end{tabular}
    }
    \label{tab:unstructured}
\end{table*}

\subsection{Exploration of Parametric Configuration of Adversarial Scratches}
\label{sec:exp_parametrization}
We explore several configurations of Adversarial Scratches, in terms of search strategy, number of scratches, order of the Bézier curve, and color configuration. All the experiments in this section are performed on $1\,000$ samples of the ImageNet dataset.

\noindent\textbf{Search Strategy:}
As discussed in Section~\ref{sec:optimizers}, we have tested four optimizers: Differential Evolution (DE)~\cite{Price2013}, Particle Swarm Optimization (PSO)~\cite{zambrano2013SPSO}, Neuro-Genetic Optimization (NGO)~\cite{nevergrad} and Random Search (RS)~\cite{zabinsky2009random}. 
We focus on the $L_0 = 400$ case, and test an attack composed of three $L_0=133$ scratches.
We test the RS optimizer using our own implementation of Sparse-RS' Random Search algorithm with scheduling. We also modify the standard DE implementation to return solutions which are within the search boundaries of Table~\ref{tab:parametrization}. Lastly, we use default PSO and NGO implementations from Nevergrad~\cite{nevergrad}. 
Table \ref{tab:optimizers} shows high fooling rates for all tested optimizers, with NGO displaying the best performance. Remarkably, these results demonstrate that Adversarial Scratches are effective across a variety of optimizers, which may be due to the small dimensionality of the parameter space required by the attack. 

\begin{table*}[t!]
    \centering
    \caption{Performance of different optimizers for the $L_0=400$ attack for three Bézier line}
    \label{tab:optimizers}
    \resizebox{0.9\textwidth}{!}{
        \begin{tabular}{|c|c|l|}
            \hline
            Optimizer & FR & Comment\\
            \hline
            RS & 83.6\% $\pm$0.5\% & Random search with step size scheduling \cite{croce2020sparse}\\
            PSO & 90.5\% $\pm$0.6\% & Implementation provided by NeverGrad \cite{nevergrad}\\
            DE & 93.9\% $\pm$0.9\% & Population size 20 and restarts each 200 iterations\\
            NGO & \textbf{97.9\% $\pm$0.3\%} & Implementation provided by NeverGrad \cite{nevergrad}\\
            \hline
        \end{tabular}
    }
\end{table*}

\begin{figure*}[t!]
    \centering
    \renewcommand\tabcolsep{0pt}
    \resizebox{1.0\textwidth}{!}{
        \begin{tabularx}{\textwidth}{p{12pt}Yp{12pt}YYYYY}
            \multirow{6}[6]{*}{\rotatebox[origin=c]{90}{\textbf{$L_0=400$}}} & Original & ~ & 1 scratch & 2 scratches & 3 scratches & 4 scratches & 5 scratches\\ 
            & 
            \setlength{\fboxsep}{0pt}\setlength{\fboxrule}{0.2pt}\fbox{\includegraphics[width=0.155\textwidth,valign=T]{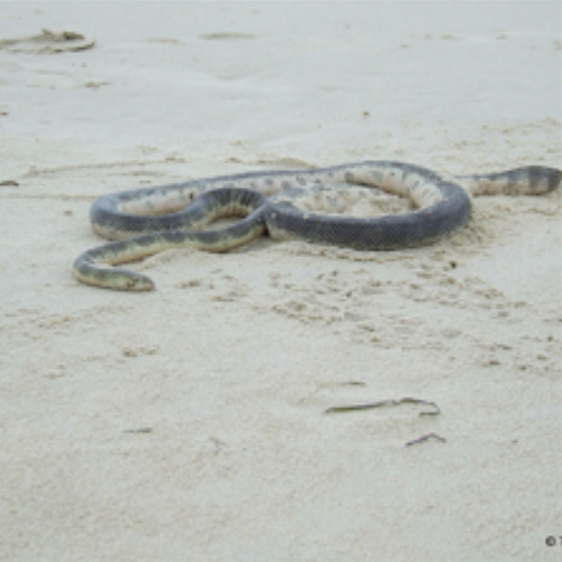}} &
            &
            \setlength{\fboxsep}{0pt}\setlength{\fboxrule}{0.2pt}\fbox{\includegraphics[width=0.155\textwidth,valign=T]{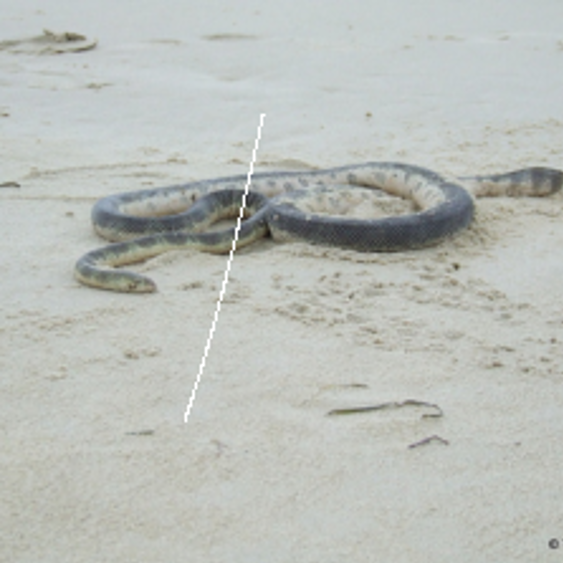}} &
            \setlength{\fboxsep}{0pt}\setlength{\fboxrule}{0.2pt}\fbox{\includegraphics[width=0.155\textwidth,valign=T]{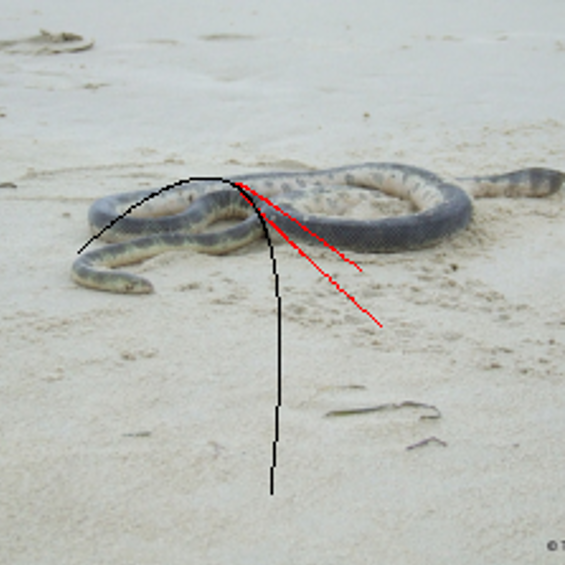}} &
            \setlength{\fboxsep}{0pt}\setlength{\fboxrule}{0.2pt}\fbox{\includegraphics[width=0.155\textwidth,valign=T]{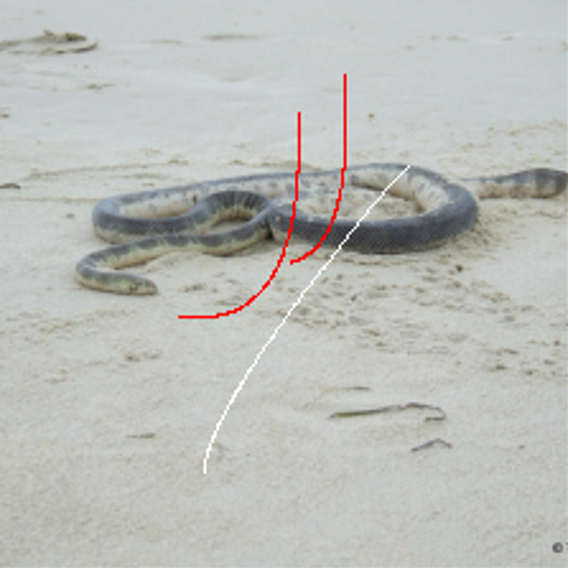}} &
            \setlength{\fboxsep}{0pt}\setlength{\fboxrule}{0.2pt}\fbox{\includegraphics[width=0.155\textwidth,valign=T]{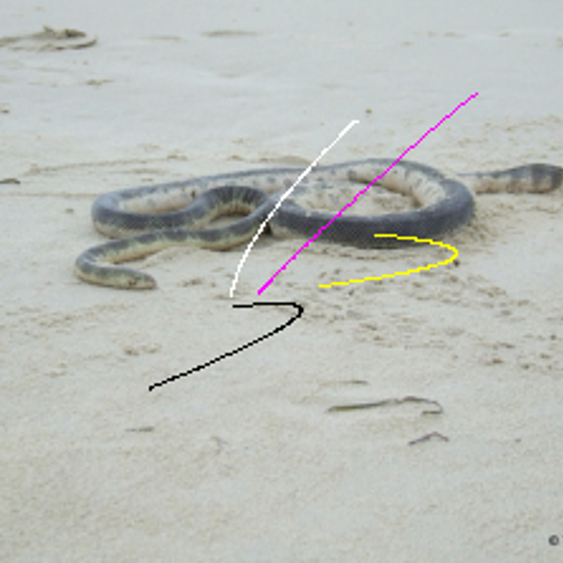}} &
            \setlength{\fboxsep}{0pt}\setlength{\fboxrule}{0.2pt}\fbox{\includegraphics[width=0.155\textwidth,valign=T]{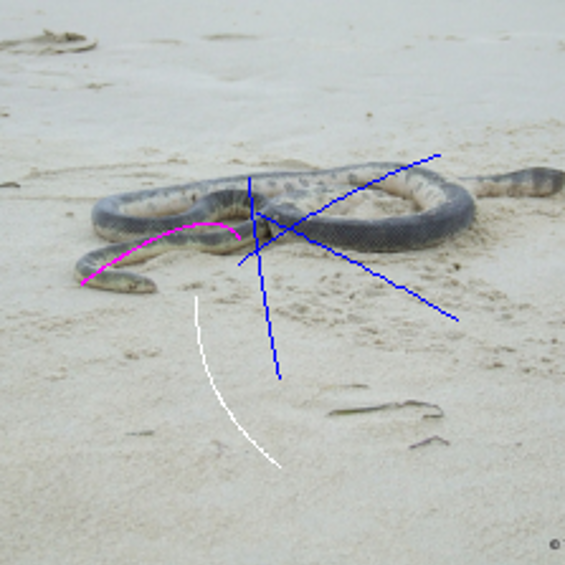}}\\
        \end{tabularx}
    }
    \caption{$L_0=400$ attacks with different number of scratches}
    \label{fig:morescratches}
\end{figure*}

\noindent\textbf{Scratch quantity:}
The number of scratches is a parameter that significantly influences deployability. As displayed in Figure~\ref{fig:morescratches}, attacks with larger number of scratches may be more powerful, since they can cover a larger region in the image. This, however, comes at the cost of deployability, as the relative position between scratches must be accounted for.

Table~\ref{tab:scratches} shows that $FR$ improves when increasing the number of scratches from $1$ to $3$, but there is no difference when further increasing from $3$ to $5$. $MQ$ decreases as the number of scratches increases, indicating a higher chance of finding a solution in the very first iterations. However, as indicated by the increase in $AQ$, search becomes more challenging as the dimensionality of the parameter space increases. 
This result supports the adoption of three scratches for the attacks discussed in Section~\ref{sec:experiments_deployable}.
From the ``Average $L_0$" column, we also deduce that using several, shorter scratches results in perturbations which are closer to the $L_0$ limit.

\begin{table*}[t!]
    \centering
    \caption{Performance of $L_0 = 400$ Bézier attacks with different number of scratches}
    \label{tab:scratches}
    \resizebox{0.9\textwidth}{!}{
        \begin{tabular}{|c|c|c|c|c|c|}
            \hline
            Bézier count & Per-Bézier $L_0$ & FR & AQ & MQ & Average $L_0$\\
            \hline
            1 & 400 & 89.6\% $\pm$0.3\% & 509 $\pm$71 & 54 $\pm$2 & \textbf{186.7 $\pm$1.7} \\
            2 & 200 & 96.6\% $\pm$0.4\% & 316 $\pm$39 & 37 $\pm$2 & 280.1.7 $\pm$0.7 \\
            3 & 133 & \textbf{97.9\% $\pm$0.3\%} & 302 $\pm$38 & 27 $\pm$3 & 331.9 $\pm$1.1 \\
            4 & 100 & \textbf{97.9\% $\pm$0.1\%} & \textbf{281 $\pm$18} & 24 $\pm$1 & 359.9 $\pm$0.6 \\
            5 & 80 & \textbf{97.9\% $\pm$0.0\%} & 301 $\pm$11 & \textbf{23 $\pm$1} & 373.2 $\pm$1.3 \\
            \hline
        \end{tabular}
    }
\end{table*}

\begin{figure*}[t!]
    \centering
    \renewcommand\tabcolsep{0pt}
    \resizebox{1.0\textwidth}{!}{
        \begin{tabularx}{\textwidth}{p{12pt}Yp{12pt}YYYY}
            \multirow{7}[7]{*}{\rotatebox[origin=c]{90}{\textbf{$L_0=400$}}} & Original & ~ & 1st order & 2nd order & 3rd order & 4th order\\ 
            & 
            \setlength{\fboxsep}{0pt}\setlength{\fboxrule}{0.2pt}\fbox{\includegraphics[width=0.186\textwidth,valign=T]{resources/more_scratches/0.png}} &
            &
            \setlength{\fboxsep}{0pt}\setlength{\fboxrule}{0.2pt}\fbox{\includegraphics[width=0.186\textwidth,valign=T]{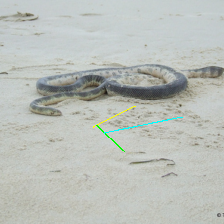}} &
            \setlength{\fboxsep}{0pt}\setlength{\fboxrule}{0.2pt}\fbox{\includegraphics[width=0.186\textwidth,valign=T]{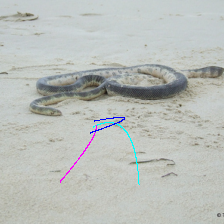}} &
            \setlength{\fboxsep}{0pt}\setlength{\fboxrule}{0.2pt}\fbox{\includegraphics[width=0.186\textwidth,valign=T]{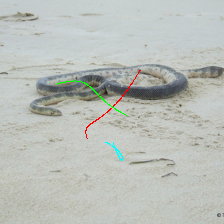}} &
            \setlength{\fboxsep}{0pt}\setlength{\fboxrule}{0.2pt}\fbox{\includegraphics[width=0.186\textwidth,valign=T]{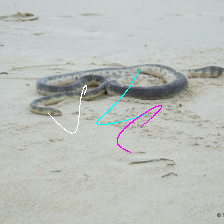}}\\
        \end{tabularx}
    }
    \caption{Example attacks with Bézier of varying order (three Bézier, each with $L_0=133$).}
    \label{fig:bezierorder}
\end{figure*}

\noindent\textbf{Order of Bézier:}
We test Adversarial Scratches modeled as Bézier curves of varying order, as displayed in Figure~\ref{fig:bezierorder}. Higher order curves may express more complex shapes, and in turn lead to better fooling rates. Such expressive power, however, comes at the cost of parametrization efficiency and deployability, since more control points need to be defined and resulting curves may be harder to draw.
The formulation~(\ref{eq:bezier}) can be generalized to express a Bézier of order $n$ as follows:
\begin{equation}
B(t) = \sum_{i=0}^{n}\binom{n}{i}(1-t)^{n-i}t^iP_i ,\, 0 \leq t \leq 1 ,
\end{equation}
where $P_i, i=0,\dots,n$ are $n+1$ control points in the form $(x_i, y_i)$. The degenerate case $n=1$ gives straight line segments, while for $n>2$ the curve may intersect itself. The parametrization of a Bézier of order $n$ requires $2(n+1) + 3$ parameters, since there are $n+1$ control points and $3$ color components.

\begin{table}[h!]
    \centering
    \caption{Performance of $L_0 = 400$ Bézier attacks with varying order}
    \label{tab:order}
    \resizebox{0.48\textwidth}{!}{
        \begin{tabular}{|c|c|c|c|}
            \hline
            Bézier order & FR & AQ & MQ\\
            \hline
            1 & 96.3\% $\pm$0.1\% & 302 $\pm$33 & 31 $\pm$3 \\
            2 & \textbf{97.9\% $\pm$0.3\%} & 302 $\pm$38 & 27 $\pm$3 \\
            3 & 97.6\% $\pm$0.2\% & 290 $\pm$24 & \textbf{25 $\pm$3} \\
            4 & 97.3\% $\pm$0.5\% & \textbf{265 $\pm$26} & 28 $\pm$2 \\
            \hline
        \end{tabular}
    }
\end{table}

We run the attack using $L_0 = 133$ Bézier curves testing Bézier order $1$ to $4$. Table~\ref{tab:order} shows that Adversarial Scratches achieve similar performance across all tested Bézier orders, with order $2$ having marginal better $FR$, order $4$ better $AQ$, and order $3$ better $MQ$. Our choice of using second-order Bézier curves in the experiments of Sections~\ref{sec:experiments_deployable} and~\ref{sec:experiments_non_deployable} is thus motivated, since they have optimal performance and are more deployable than higher-order ones.

\begin{figure*}[t!]
    \centering
    \renewcommand\tabcolsep{0pt}
    \resizebox{1.0\textwidth}{!}{
        \begin{tabularx}{\textwidth}{p{12pt}Yp{12pt}YYYY}
            \multirow{10}[10]{*}{\rotatebox[origin=c]{90}{\textbf{$L_0=400$}}} & Original & ~ & monochrome saturated & polychrome gray-scale & polychrome image-color\\ 
            & 
            \setlength{\fboxsep}{0pt}\setlength{\fboxrule}{0.2pt}\fbox{\includegraphics[width=0.232\textwidth,valign=T]{resources/more_scratches/0.png}} &
            &
            \setlength{\fboxsep}{0pt}\setlength{\fboxrule}{0.2pt}\fbox{\includegraphics[width=0.232\textwidth,valign=T]{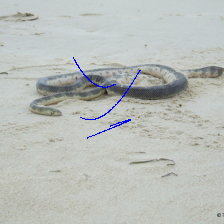}} &
            \setlength{\fboxsep}{0pt}\setlength{\fboxrule}{0.2pt}\fbox{\includegraphics[width=0.232\textwidth,valign=T]{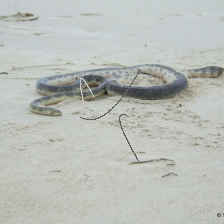}} &
            \setlength{\fboxsep}{0pt}\setlength{\fboxrule}{0.2pt}\fbox{\includegraphics[width=0.232\textwidth,valign=T]{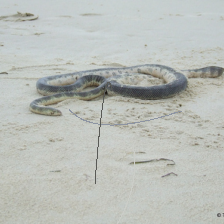}}\\
        \end{tabularx}
    }
    \caption{Example attacks with different color parametrization (three Bézier, each with $L_0=133$).}
    \label{fig:beziercolor}
\end{figure*}

\noindent\textbf{Color Configuration:}
We test different color configurations and analyze changes in attack performance. Focusing on an attack composed of three quadratic $L_0 = 133$ Bézier curves, we test:
\begin{itemize}[leftmargin=*]
    \item A ``polychrome, saturated" attack where each scratch assumes one of eight fully saturated colors;
    \item A ``monochrome, saturated" attack where all scratches have the same fully saturated color;
    \item A ``polychrome, gray-scale" attack where each scratch assumes one gray-scale color;
    \item A ``polychrome, image-color" attack where each scratch assumes color equal to one of the available pixels in the image.
\end{itemize}

\begin{table}[t!]
    \centering
    \caption{Performance of $L_0 = 400$ Bézier attacks with varying color configuration}
    \label{tab:color}
    \resizebox{0.48\textwidth}{!}{
        \begin{tabular}{|c|c|c|c|}
            \hline
            Color configuration & FR & AQ & MQ\\
            \hline
            polychrome, saturated & \textbf{97.9\% $\pm$0.3\%} & 302 $\pm$38 & 27 $\pm$3 \\
            monochrome, saturated & 97.6\% $\pm$0.1\% & \textbf{252 $\pm$24} & \textbf{24} $\pm$2 \\
            polychrome, gray-scale & 92.4\% $\pm$0.6\% & 534 $\pm$27 & 94 $\pm$4 \\
            polychrome, image-color & 87.7\% $\pm$0.6\% & 757 $\pm$66 & 128 $\pm$9 \\
            \hline
        \end{tabular}
    }
\end{table}
Figure~\ref{fig:beziercolor} shows example scratches with these color configurations. Table \ref{tab:color} shows that the ``monochrome, saturated" attack has slightly better performance in terms of AQ and MQ than the baseline attack with three independently colored scratches, while still matching it in terms of FR. The gray-scale and image-color attack also show good performance, albeit lower than that of the other attacks, as less saturated colors have diminished attacking power.

\section{Discussion}
\label{sec:discussion}
Our experiments show that Adversarial Scratches outperform other state-of-the-art deployable attacks, achieving comparable performance even against non-deployable $L_0$ black-box attacks.
Furthermore, our experiments on the TSRD dataset show promising results for the applicability of Adversarial Scratches to physical targets.
We attribute the success of Adversarial Scratches to the greatly reduced search space compared to other attacks, especially those presented in Sparse-RS. Indeed, the ``any-pixel'' attack has a search space with dimensionality $5k$, where $k$ is the $L_0$ bound. Adversarial Scratches, instead, only require $9$ parameters for each scratch, independently of the $L_0$ bound. In our experiments, we used single scratches for the $L_0 = 50$ case, which means a reduction of parameter count of more than $27\times$ ($9$ parameters for a single scratch attack against $250$ parameters for $L_0=50$ ``any-pixel'' attack). For the $L_0 = 400$, we used $3$ scratches, resulting in $74$ times fewer parameters ($27$ parameters for three scratches against $2000$ parameters for $L_0=400$ ``any-pixel'' attack).

\subsection{Defenses}

\begin{figure}[t!]
    \centering
    \renewcommand\tabcolsep{0pt}
    \resizebox{0.48\textwidth}{!}{
        \begin{tabularx}{\textwidth}{Yp{10pt}YYY}
            Adversarial & & Adversarial & JPEG compressed & Median filtered \\ 
            \multicolumn{5}{c}{\includegraphics[width=\textwidth,valign=T]{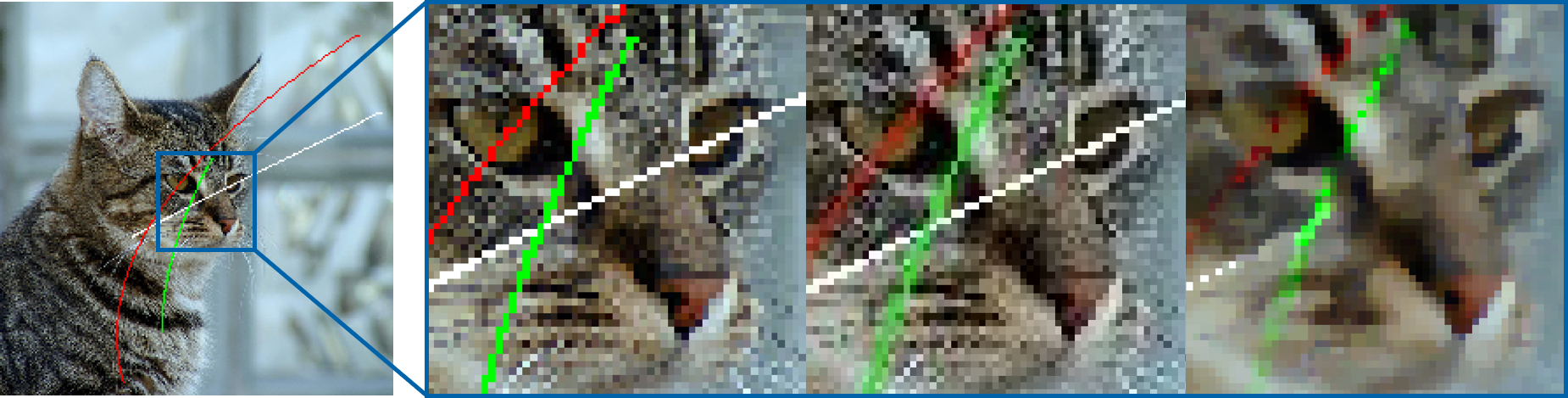}}\\
        \end{tabularx}
    }

    \caption{Defenses to Adversarial Scratches. We visualize the results of two defense mechanisms against three Adversarial Scratches, each with $L_0$ bound $133$. The defenses are JPEG compression and median filtering. Median filtering is more effective as it removes most of the scratches from the image, while JPEG compression only marginally alters the perturbation.}
    \label{fig:img_domain_compression}
\end{figure}

Recent works have proposed methodologies to counteract the effects of adversarial perturbations. These techniques encompass robust training procedures~\cite{li2022semi}, perturbation detection and removal~\cite{han20222}, and reconstructions through deep image priors~\cite{dai2022deep}. To defend from Adversarial Scratches, following~\cite{evtimov2020security}, we consider defenses that rely on \emph{input filtering}, as these are more scalable than defenses that try to make the model itself more robust. In particular, we adopt:
\begin{itemize}[leftmargin=*]
    \item \textbf{JPEG compression} with varying quality factors.
    \item \textbf{Median filtering} separately to each channel with a kernel size of $3\times3$ pixels.
\end{itemize}

\noindent\textbf{Metrics:}
We assess the effectiveness of defense $d(\cdot)$ through the \emph{recovery rate} ($RR$), defined as the fraction of successful adversarial samples whose filtered version is correctly classified:
\begin{equation}
    \label{eq:recoveryrate}
    RR = \frac{|\{C(\mathbf{x}) = y\} \bigcap \{C(\mathbf{x}')\neq y\} \bigcap \{C(d(\mathbf{x}'))=y\}|}{|\{C(\mathbf{x}) = y\} \bigcap \{C(\mathbf{x}')\neq y\}|},
\end{equation}
where $X' = \{\mathbf{x}'_1, ..., \mathbf{x}'_n\}$ is the set of perturbed images (possibly including those for which the attack was not successful) for samples $X = \{\mathbf{x}_1, ..., \mathbf{x}_n\}$.\footnote{If the filtered image is classified differently than the adversarial one $C(\mathbf{x}') \neq C(d(\mathbf{x}))$, but the resulting class is not the target $C(d(\mathbf{x})) \neq y$, the image is not considered to be recovered.}

\noindent\textbf{Results:}
We assess the effectiveness of these defenses against a ``polychrome, saturated" attack, using three $L_0 = 133$ second-order Bézier curves. The attack achieved $97.9\%$ $FR$ on 1000 samples of ImageNet, therefore, our defense analysis is based on a large pool of 979 adversarial samples, from which we exclude originally misclassified images (\ref{eq:recoveryrate}).

\begin{table*}[t!]
    \centering
    \caption{Recovery Rate and effects of the defenses on model performance.}
    \label{tab:recoveryrate}
    \resizebox{0.9\textwidth}{!}{
        \renewcommand{\arraystretch}{1.2}
        \begin{tabular}{|c|c|c|c|}
            \hline
            \multirow{2}*{Defense strategy} & Adversarial samples & \multicolumn{2}{c|}{Original images} \\
            \cline{2-4}
            & Recovery Rate & Model performance & Performance delta \\
            \hline
            \renewcommand{\arraystretch}{1}
            \emph{No defense (baseline)} & \emph{0\%} & \textbf{\emph{76.7\%}} & \textbf{\emph{0\%}}\\
            Median filtering, $3 \times 3$ kernel & \textbf{77.1}\% & 73.0\% & -3.7\%\\
            JPEG, quality = 85 & 41.4\% & 74.4\% & -2.3\%\\
            JPEG, quality = 90 & 39.0\% & 74.9\% & -1.8\%\\
            JPEG, quality = 95 & 32.6\% & 75.7\% & -1.0\%\\
            JPEG, quality = 99 & 25.7\% & 76.1\% & -0.6\%\\
            \hline
        \end{tabular}
    }
\end{table*}

Table~\ref{tab:recoveryrate} shows that both JPEG compression and median filtering are effective defenses, with median filtering providing the highest recovery rate ($77.1\%$). Figure~\ref{fig:img_domain_compression} shows that median filtering can indeed remove scratch pixels from the image. JPEG compression is also a viable technique to recover the original target class (up to $41.4\%$ recovery rate with quality $=85$).

A drawback of filtering-based defenses is that these transformations may result in a drop in classification performance on non-adversarial images. To study this phenomenon, we compute the performance of the CNN used for our experiments on 1000 samples from the ImageNet validation set. Then, we apply the defense to each image in this set, and compute the performance drop on the filtered samples. Table~\ref{tab:recoveryrate} shows that the most effective median filtering is also the most detrimental to the model's performance, resulting in a drop in accuracy of $3.7\%$. We argue, however, that the high recovery rate justifies this minor performance drop.

\subsection{Targeted Attacks}
\label{sec:targetedattacks}
In our work, we have focused on untargeted attacks. Nonetheless, Adversarial Scratches are easily extendable to the targeted scenario. As detailed in Section \ref{sec:categorization}, targeted attacks generate adversarial samples $\mathbf{x}'$ that are classified as belonging to a specific target class $y'$, namely $C(\mathbf{x}') = y'$.
To obtain targeted Adversarial Scratches, we replace the optimization objective (\ref{eq:margin_loss}) with a cross entropy loss $\mathcal{H}$, where the target vector $\mathbbm{1}_{y'}$ is all zeroes except for a one in position $y'$:

\begin{equation}
    \small
    \label{eq:cross_entropy_loss}
        \mathcal{H}(f(x),\mathbbm{1}_{y'}) = -\log(f(x)_{y'}) - \sum_{i\neq y'}\log(1-f(x)_i)\,.
\end{equation}

In the untargeted scenario, the lower bound $\mathcal{L}_f(\mathbf{x}, \mathbf{x}')$ (\ref{eq:margin_loss_target}) identifies the threshold below which an image is adversarial. This is not the case for targeted attacks, where classification (and thus attack success) must be explicitly verified by checking whether $C(\mathbf{x}') = y'$. 

\section{Conclusions and Future Works}
\label{sec:conclusion}
In this paper, we propose \emph{Adversarial Scratches}: a novel attack structured as parametric Bézier curves applied to the image, and designed to be \emph{deployable} to physical targets. We believe that our study of adversarial attacks is very relevant to the security of applications making use of deep learning models, which must be robust to attacks, and especially so to those attacks that can be deployed in the real world. On the one hand, our study demonstrates that Adversarial Scratches are effective in a variety of scenarios, including attacks against a publicly available API, even though it requires modification of very few pixels. On the other hand, we have presented filtering-based countermeasures to mitigate the vulnerabilities originating from Adversarial Scratches, and have quantitatively assessed the impact of these defenses on the model's performance.

In this work, we have limited our scope to one pixel wide Bézier curves targeting image classifiers. Future works may on the one hand study the effects of such attacks against models addressing higher visual recognition tasks, such as object detection and segmentation. On the other hand, future works may expand on the concept of Adversarial Scratches, proposing deployable attacks based on different parametric models. We also plan on developing countermeasures to extended versions of Adversarial Scratches, as the proposed filtering techniques may not be effective.
Another important direction is that of studying the robustness of Adversarial Scratches to realistic changes in the image acquisition, including pose, light conditions, and background contents. In this work, we focused on applying scratches to single views of traffic signs, however, attacks that are successful regardless of acquisition settings may pose very serious threats to critical systems in the real-world, such as autonomous vehicles and AI-powered security cameras.
Also in this case, we plan to develop new countermeasures, as such robust attacks may not be affected by simple filtering.

\section{Acknowledgements}
This research did not receive any specific grant from funding agencies in the public, commercial, or not-for-profit sectors. We gratefully acknowledge the support of NVIDIA for the four A6000 GPUs granted through the Applied Research Accelerator Program to Politecnico di Milano.

\bibliographystyle{ACM-Reference-Format}
\bibliography{main}

\end{document}